\documentclass{article}

    \PassOptionsToPackage{compress}{natbib}
\usepackage[final]{neurips_2023}

\usepackage[utf8]{inputenc} % allow utf-8 input
\usepackage[T1]{fontenc}    % use 8-bit T1 fonts
\usepackage{hyperref}       % hyperlinks
\usepackage{url}            % simple URL typesetting
\usepackage{booktabs}       % professional-quality tables
\usepackage{amsfonts}       % blackboard math symbols
\usepackage{nicefrac}       % compact symbols for 1/2, etc.
\usepackage{microtype}      % microtypography
\usepackage{xcolor}         % colors

\usepackage{graphicx}
\usepackage{caption}
\usepackage{subcaption}
\usepackage{wrapfig}

\usepackage{algorithm}
\usepackage{algpseudocode}
\usepackage{amsmath}

\newcommand{\ent}[0]{\mathcal{H}}
\newcommand{\norm}[1]{\left\lVert#1\right\rVert}

\DeclareRobustCommand{\rchi}{{\mathpalette\irchi\relax}}
\newcommand{\irchi}[2]{\raisebox{\depth}{$#1\chi$}} 

\def\ppo{PPO}

\def\ppg{PPG}

\def\ucbdrac{UCB-DrAC}
\def\plr{PLR}

\def\daac{DAAC}
\def\idaac{IDAAC} 
\def\leep{LEEP} 

\newcommand{\methodname}{Explore to Generalize}
\newcommand{\methodnameshort}{ExpGen}

\title{Explore to Generalize in Zero-Shot RL}

\author{%
  Ev Zisselman\thanks{Correspondence E-mail: \texttt{ev\textunderscore zis@campus.technion.ac.il}},~ Itai Lavie, ~Daniel Soudry, ~Aviv Tamar\\
  Technion -- Israel Institute of Technology
}

\begin{document}

\maketitle

\begin{abstract}
We study zero-shot generalization in reinforcement learning---optimizing a policy on a set of training tasks to perform well on a similar but unseen test task. 
To mitigate overfitting, previous work explored different notions of invariance to the task. However, on problems such as the ProcGen Maze, an adequate solution that is invariant to the task visualization does not exist, and therefore invariance-based approaches fail. 
Our insight is that learning a policy that effectively \textit{explores} the domain is harder to memorize than a policy that maximizes reward for a specific task, and therefore we expect such learned behavior to generalize well; we indeed demonstrate this empirically on several domains that are difficult for invariance-based approaches. Our \textit{{\methodname}} algorithm (\methodnameshort) builds on this insight: we train an additional ensemble of agents that optimize reward. At test time, either the ensemble agrees on an action, and we generalize well, or we take exploratory actions, which generalize well and drive us to a novel part of the state space, where the ensemble may potentially agree again. We show that our approach is the state-of-the-art on tasks of the ProcGen challenge that have thus far eluded effective generalization, yielding a success rate of $83\%$ on the Maze task and $74\%$ on Heist with $200$ training levels. \methodnameshort\ can also be combined with an invariance based approach to gain the best of both worlds, setting new state-of-the-art results on ProcGen.
Code available at \url{https://github.com/EvZissel/expgen}. 
\end{abstract}

\section{Introduction}
\label{Introduction}
\begin{figure}[t]
    \centering
     \begin{subfigure}[h]{0.15\textwidth}
        \centering
         \includegraphics[width=\textwidth]{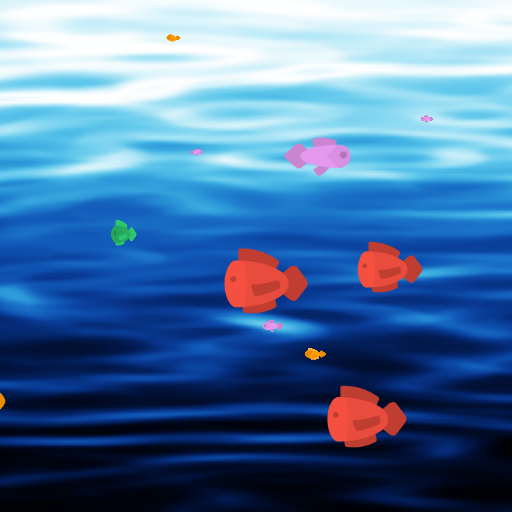}
          \caption{BigFish}
         \label{bigfish}
     \end{subfigure}
     \begin{subfigure}[h]{0.15\textwidth}
        \centering
         \includegraphics[width=\textwidth]{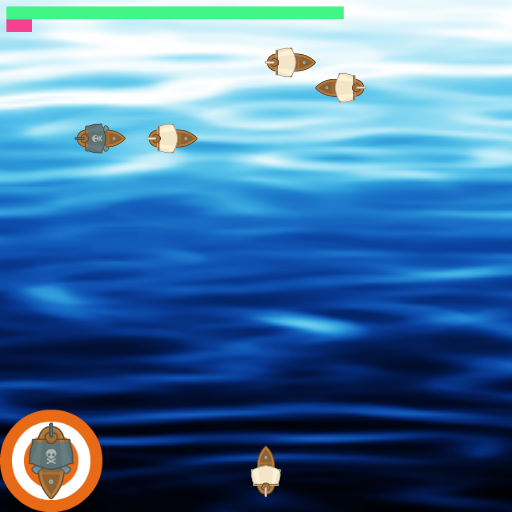}
          \caption{Plunder}
         \label{plunder}
     \end{subfigure}
     \begin{subfigure}[h]{0.15\textwidth}
        \centering
         \includegraphics[width=\textwidth]{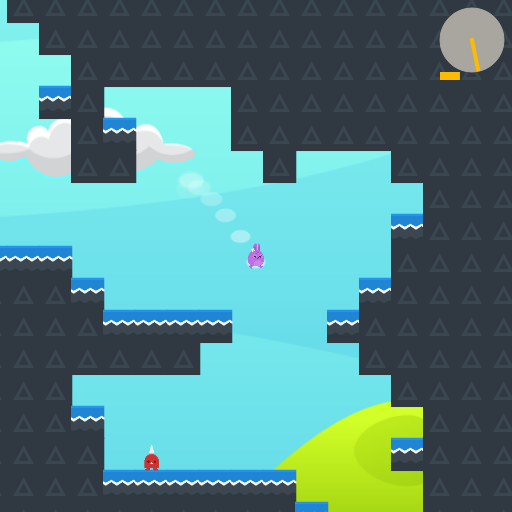}
          \caption{Jumper}
         \label{jumper}
     \end{subfigure}
     \begin{subfigure}[h]{0.15\textwidth}
        \centering
         \includegraphics[width=\textwidth]{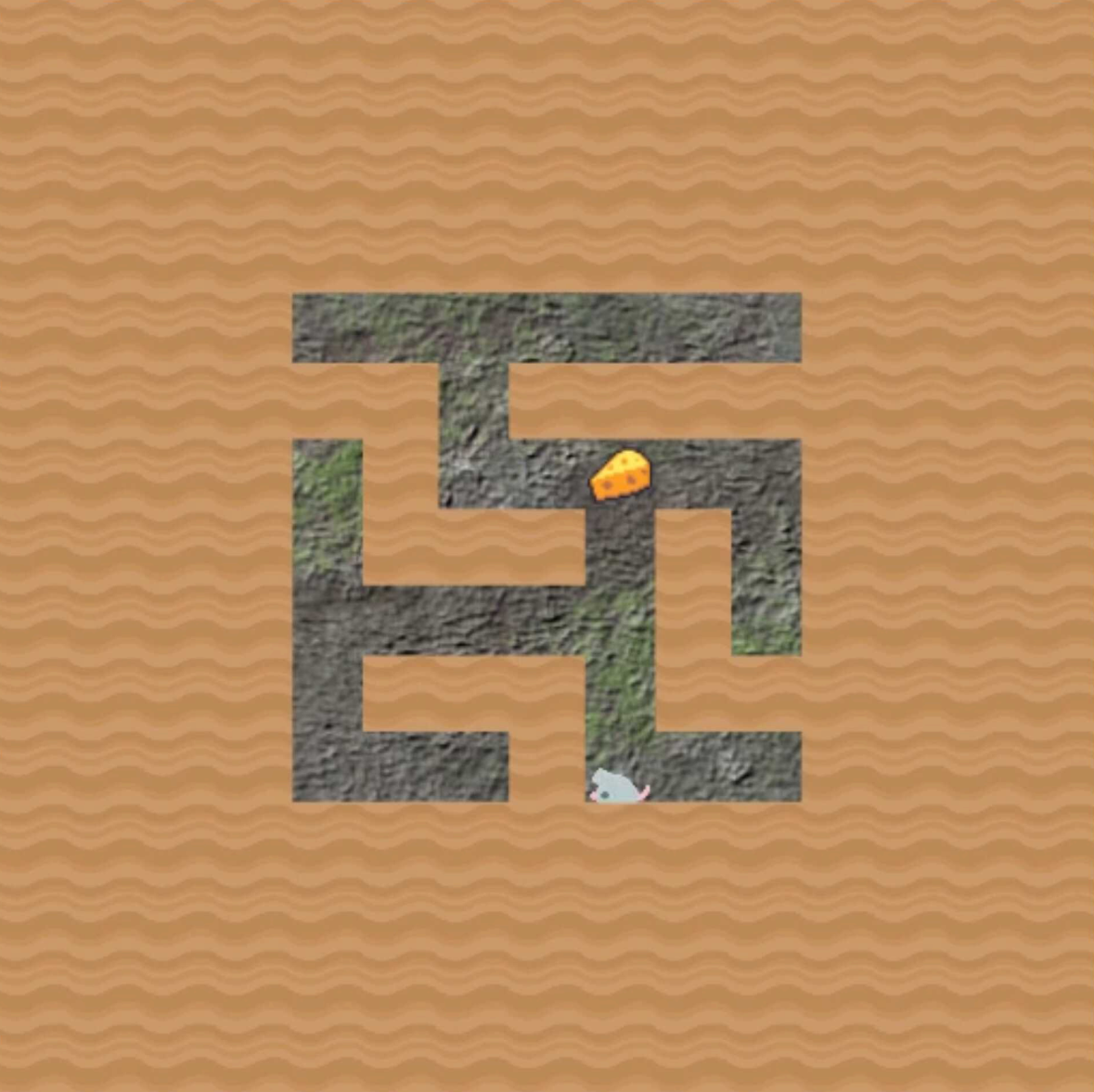}
          \caption{Maze}
         \label{maze}
     \end{subfigure}
     \begin{subfigure}[h]{0.15\textwidth}
        \centering
         \includegraphics[width=\textwidth]{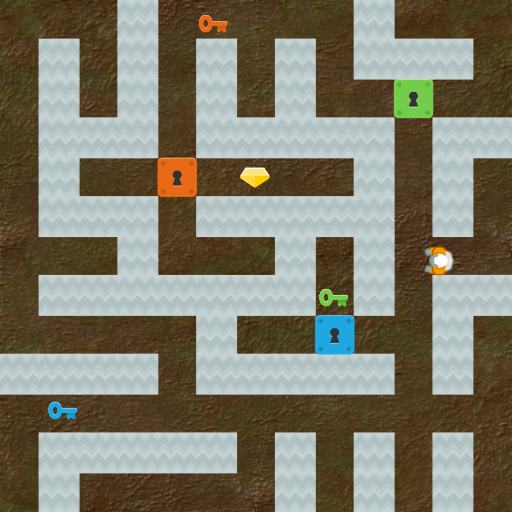}
          \caption{Heist}
         \label{heist}
     \end{subfigure}
     \hfill
    \begin{subfigure}[h]{0.15\textwidth}
        \centering
         \includegraphics[width=\textwidth]{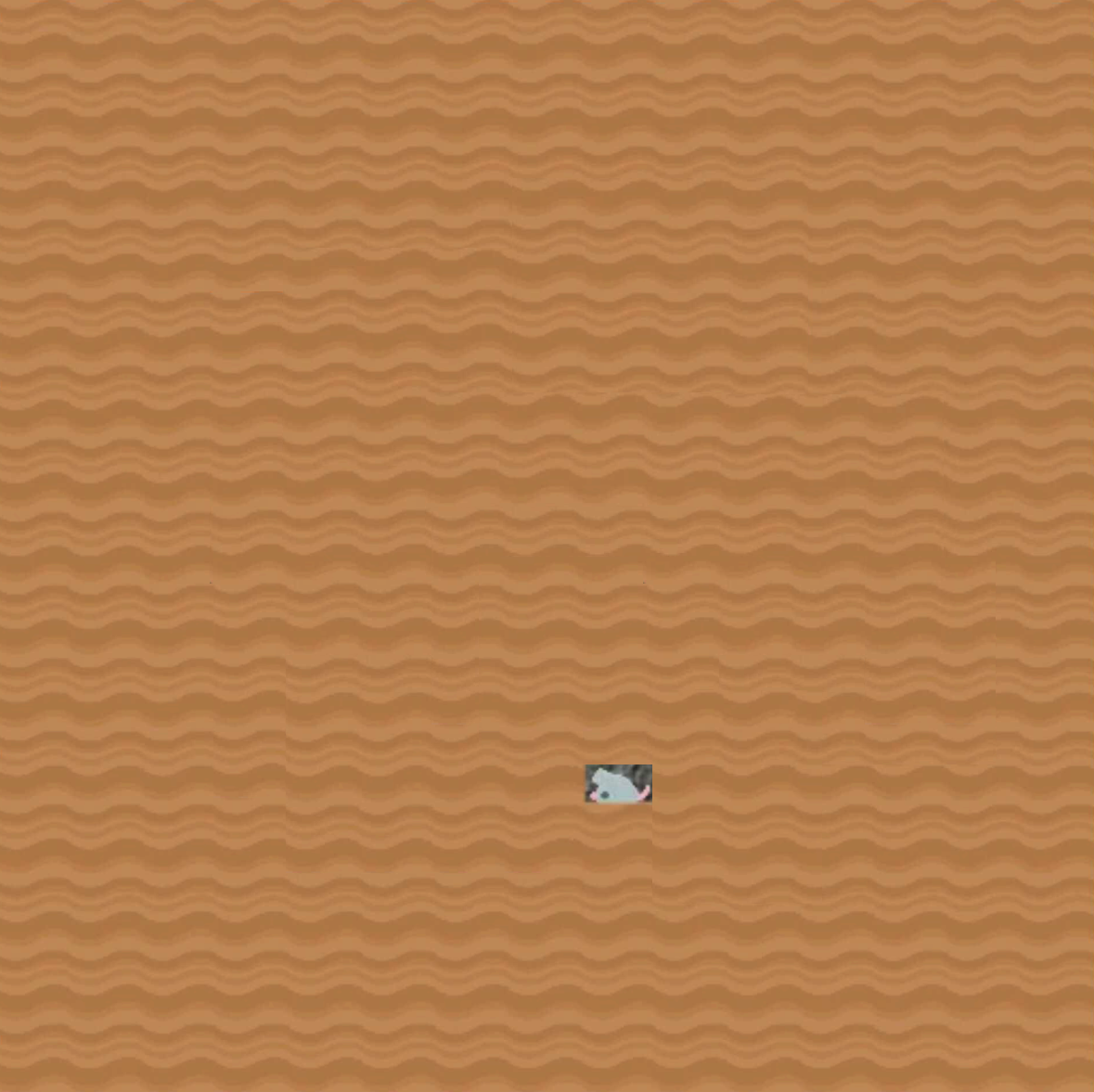}
          \caption{Hidden Maze}
         \label{hidden}
     \end{subfigure}
    \caption{(a),(b),(c),(d) and (e) displays screenshot of ProcGen games. (f) Imaginary maze with goal and walls removed (see text for explanation).
    } 
    \label{fig:environments}
\end{figure}

Recent developments in reinforcement learning (RL) led to algorithms that surpass human experts in a broad range of tasks \citep{mnih2015human,vinyals2019grandmaster,schrittwieser2020mastering,wurman2022outracing}. 
In most cases, the RL agent is tested on the same task it was trained on, and is not guaranteed to perform well on unseen tasks.
In zero-shot generalization for RL (ZSG-RL), however, the goal is to train an agent on training domains to act optimally in a new, previously unseen test environment~\citep{kirk2021survey}. A standard evaluation suite for ZSG-RL is the ProcGen benchmark~\citep{cobbe2020leveraging}, containing 16 games, each with levels that are procedurally generated to vary in visual properties (e.g., color of agents in BigFish, Fig.~\ref{bigfish}, or background image in Jumper, Fig.~\ref{jumper}) and dynamics (e.g., wall positions in Maze, Fig.~\ref{maze}, and key positions in Heist, Fig.~\ref{heist}). 

Previous studies focused on identifying various \textit{invariance properties} in the tasks, and designing corresponding \textit{invariant policies}, through an assortment of regularization and augmentation techniques \citep{igl2019generalization,cobbe2019quantifying,wang2020improving,lee2019network,raileanu2021automatic,raileanu2021decoupling, cobbe2021phasic,sonar2021invariant,bertran2020instance,li2021domain}. For example, a policy that is invariant to the color of agents is likely to generalize well in BigFish. More intricate invariances include the order of observations in a trajectory~\citep{raileanu2021decoupling}, and the length of a trajectory, as reflected in the value function~\citep{raileanu2021decoupling}.

\begin{wrapfigure}{r}{7.0cm}
\includegraphics[trim={0 0 0 0cm},clip, width=7.0cm]{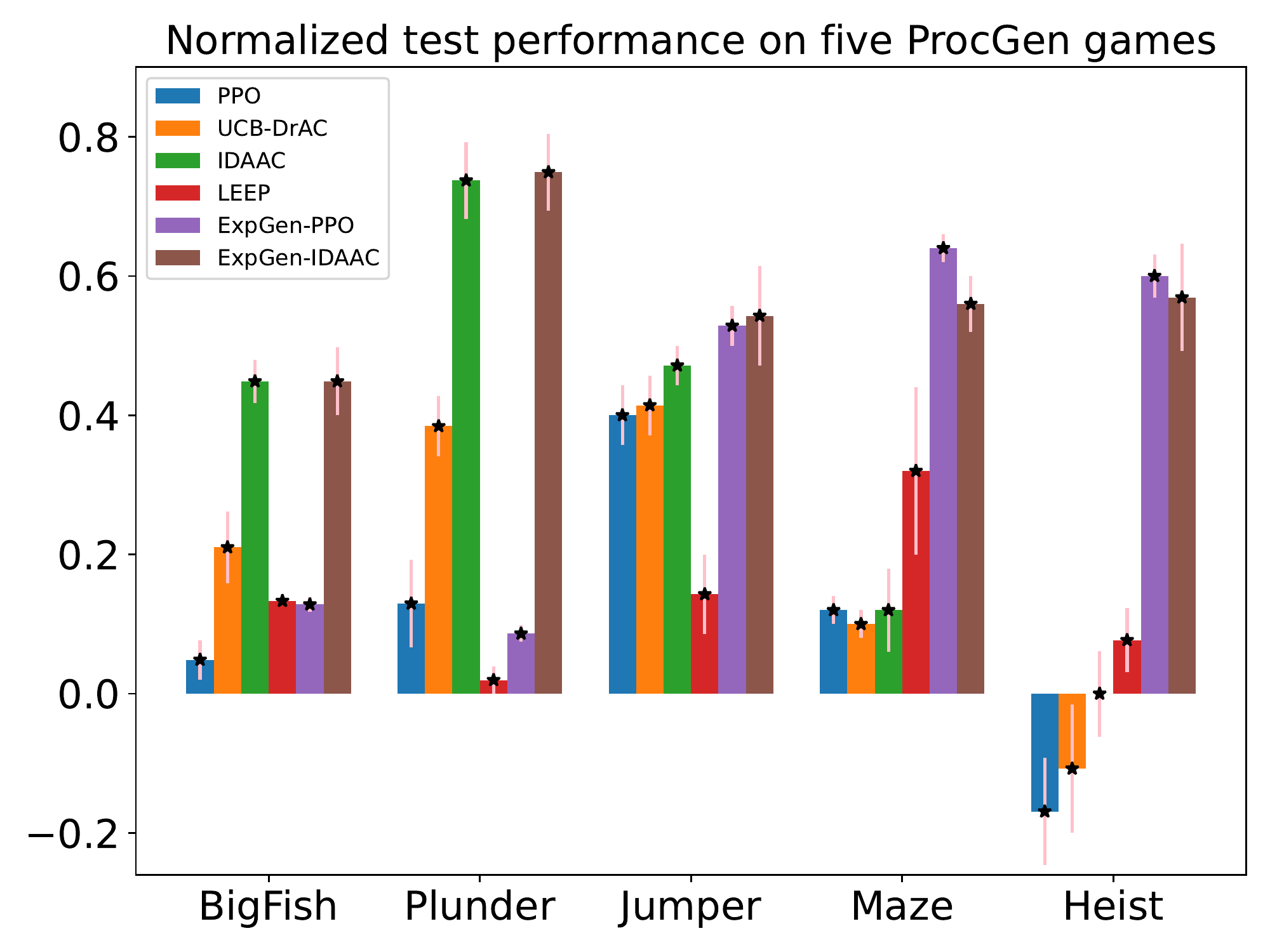}
\setlength{\abovecaptionskip}{-.3cm}
\setlength{\belowcaptionskip}{-.4cm}
\caption{
Normalized test Performance for \methodnameshort, \leep{}, \idaac{}, \daac{}, and \ppo{}, on five ProcGen games. ExpGen shows state-of-the-art performance on test levels of Maze, Heist and Jumper; games that are notoriously challenging for other leading approaches. The scores are normalized as proposed by \citep{cobbe2020leveraging}.}

\label{fig:Test_performance_on_Procgen}
\end{wrapfigure}

Can ZSG-RL be reduced to only finding invariant policies? As a counter-argument, consider the following thought experiment\footnote{We validated this experiment empirically, using a recurrent policy on Maze with 128 training tasks, observing 85\% success rate on training domains, and test success similar to a random policy (see Appendix \ref{appendix:hidden_maze_exp}).}. Imagine Maze, but with the walls and goal hidden in the observation (Fig. \ref{hidden}). Arguably, this is the most task-invariant observation possible, such that a solution can still be obtained in a reasonable time.
An agent with memory can be trained to optimally solve all training tasks: figuring out wall positions by trying to move ahead and observing the resulting motion, and identifying based on its movement history in which training maze it is currently in. Obviously, such a strategy will not generalize to test mazes. Indeed, as depicted in Figure \ref{fig:Test_performance_on_Procgen}, performance in tasks like Maze and Heist, where the strategy for solving any particular training task must be \textit{indicative} of that task, has largely not improved by methods based on invariance (e.g. \ucbdrac{} and \idaac{}). 

Interestingly, decent zero-shot generalization can be obtained even without a policy that generalizes well. As described by \citet{ghosh2021generalization}, an agent can overcome test-time errors in its policy by treating the perfect policy as an \textit{unobserved} variable. The resulting decision making problem, termed the \textit{epistemic POMDP}, may require some exploration at test time to resolve uncertainty. \citet{ghosh2021generalization} further proposed the LEEP algorithm based on this principle, which trains an ensemble of agents and essentially chooses randomly between the members when the ensemble does not agree, and was the first method to present substantial generalization improvement on Maze.  

In this work, we follow the epistemic POMDP idea, but ask: \textit{how to improve exploration at test time?} Our approach is based on a novel discovery: when we train an agent to \textit{explore} the training domains using a maximum entropy objective~\citep{hazan2019provably, mutti2021task}, we observe that the learned exploration behavior generalizes surprisingly well---much better than the generalization attained when training the agent to maximize reward. Intuitively, this can be explained by the fact that reward is a strong signal that leads to a specific behavior that the agent can `memorize' during training, while exploration is naturally more varied, making it harder to memorize and overfit. 

Exploration by itself, however, is not useful for solving new tasks. Our algorithm, \textit{\methodname} (\methodnameshort), additionally trains an ensemble of reward-seeking agents. At test time, either the ensemble agrees on an action, and we generalize well, or we take exploratory actions \textit{using the exploration policy}, which we demonstrate to generalize, and drive us to a novel part of the state space, where the ensemble may potentially agree again.

\methodnameshort\ is simple to implement, and can be used with any reward maximizing RL algorithm. Combined with vanilla PPO, \methodnameshort\ significantly improves the state-of-the-art (SOTA) on several ProcGen games for which previous methods fail (see Fig.~\ref{fig:Test_performance_on_Procgen}). \methodnameshort\ also significantly improves upon LEEP, due to its effective test-time exploration strategy.
For example, on Maze with $200$ training levels, our method obtains $83\%$ success on test tasks, whereas the previous state-of-the-art achieved $66\%$. When combined with \idaac{}~\citep{raileanu2021decoupling}, the leading invariance-based algorithm, \methodnameshort\ achieves state-of-the-art performance on the full ProcGen suite (the full results are provided in Appendix~\ref{appendix:results_all_idaac_procgen}).

\section{Related Work}
\label{Related_Work}
\paragraph{Generalization in RL} The recent survey by \citet{kirk2021survey} provides an extensive review of generalization in RL; here, we provide a brief overview.
One approach to generalization is by artificially increasing the number of training tasks, using either procedural generation~\citep{cobbe2019quantifying, cobbe2020leveraging}, or augmentations~\citep{kostrikov2020image, ye2020rotation, lee2019network, raileanu2021automatic}, task interpolation~\citep{yao2021meta} or various regularization technique, such as dropout~\citep{igl2020impact} and batch normalization~\citep{farebrother2018generalization, igl2020impact}. {Leading approaches, namely \idaac{}
\citep{raileanu2021decoupling} and \ppg{}~\citep{cobbe2021phasic}, investigate the advantages of decoupling policy and value functions for generalization, whereas} \citet{jiang2021prioritized} propose automatic curriculum learning of levels.

A different approach is to add inductive bias to the neural network policy or learning algorithm. Approaches such as \citet{tamar2016value,vlastelica2021neuro,boutilier2020differentiable} embed a differentiable planner or learning algorithm into the neural network. Other methods \citep{kansky2017schema,toyer2018action,rivlin2020generalized} combine learning with classical graph planning to generalize across various planning domains. These approaches require some knowledge about the problem structure (e.g., a relevant planning algorithm), while our approach does not require any task-specific knowledge. 
Another line of work aims to learn policies or features that are invariant across the different training tasks~\citep{sonar2021invariant,bertran2020instance,li2021domain, igl2019generalization, stooke2021decoupling, mazoure2020deep} and are thus more robust to sensory variations.

% Interestingly, earlier works that employ ensembles as a way to balance exploration and exploitation show potential. 
Using an ensemble to direct exploration to unknown areas of the state space was proposed in the model-based TEXPLORE algorithm of \citet{hester2013texplore}, where an ensemble of transition models was averaged to induce exploratory actions when the models differ in their prediction.
% Here, an aggregate tree model forms the ensemble's prediction, such that a broad agreement between the individual models promotes exploitation. However, when the constituent models differ in their predictions of the next state, the aggregate model becomes more stochastic, thus encouraging exploration. To illustrate, an ensemble comprising $m$ decision trees, each predicting a different next state, yields an aggregate model that induces a uniform distribution over the $m$ possible next states. In this case, TEXPlore uses random exploration in the state-action space.
Observing that exploration can help with zero-shot generalization, the model-free \leep{}~algorithm by \citet{ghosh2021generalization} is most relevant to our work. \leep{}~trains an ensemble of policies, each on a separate subset of the training environment, with a loss function that encourages agreement between the ensemble members. Effectively, the KL loss in \leep{}~encourages random actions when the agents of the ensemble do not agree, which is related to our method. However, random actions can be significantly less effective in exploring a domain than a policy that is explicitly trained to explore, such as a maximum-entropy policy. Consequentially, we observe that our approach leads to significantly better performance at test time.

\paragraph{State Space Maximum Entropy Exploration}
Maximum entropy exploration (maxEnt, \citealt{hazan2019provably, mutti2021task}) is an unsupervised learning framework
that trains policies that maximize the entropy of their state-visitation frequency, leading to a behavior that continuously explores the environment state space. Recently, maximum entropy policies have gained attention in RL \citep{liu2021behavior, liu2021aps, yarats2021reinforcement, seo2021state, hazan2019provably, mutti2021task} mainly in the context of unsupervised pre-training. In that setting, the agent is allowed to train for a long period without access to environment rewards, and only during test the agent gets exposed to the reward signal and performs a limited fine-tuning adaptation learning. Importantly, these works expose the agent to the same environments during pre-training and test phases, with the only distinction being the lack of extrinsic reward during pre-training. 
To the best of our knowledge, our observation that maxEnt policies generalize well in the zero-shot setting is novel.

\section{Problem Setting and Background}
\label{sec:Problem_Setting}
We describe our problem setting and provide background on maxEnt exploration.
\subsection*{Reinforcement Learning (RL)}
In Reinforcement Learning an agent interacts with an unknown, stochastic environment and collects rewards. This is modeled by a Partially Observed Markov Decision Process (POMDP) \citep{bertsekas2012dynamic}, which is the tuple $M = (S, A, O, P_{init}, P, \Sigma, r, \gamma)$, where $S \in \mathbb{R}^{|S|}$ and $A \in \mathbb{R}^{|A|}$ are the state and actions spaces, $O$ is the observation space, $P_{init}$ is an initial state distribution, $P$ is the transition kernel, $\Sigma$ is the observation function, $r : S \times A \rightarrow \mathbb{R}$ is the reward function, and $\gamma \in [0,1)$ is the discount factor. The agent starts from initial state $s_0 \sim P_{init}$ and at time $t$ performs an action $a_t$ on the environment that yields a reward $r_t = r(s_t,a_t)$, and an observation $o_t = \Sigma(s_t,a_t) \in O$. Consequently, the environment transitions into the next state according to $s_{t+1} \sim P(\cdot| s_t,a_t)$. Let the history at time $t$ be $h_t = \{o_0, a_0, r_0, o_1, a_1, r_1 \ldots, o_t\}$, the sequence of observations, actions and rewards. The agent's next action is outlined by a policy $\pi$, which is a stochastic mapping from the history to an action probability
$\pi(a|h_t) = P(a_t = a|h_t)$. {In our formulation,} a history-dependent policy (and not a Markov policy) is required both due to partially observed states, epistemic uncertainty~\citep{ghosh2021generalization}, and also for optimal maxEnt exploration~\citep{mutti2022importance}. 

\subsection*{Zero-Shot Generalization for RL}

We assume a prior distribution over POMDPs $P(M)$, defined over some space of POMDPs. For a given POMDP, an optimal policy maximizes the expected discounted return $\mathbb{E}_{\pi,M}[ \sum_{t=0}^\infty \gamma^tr(s_t,a_t)]$, where the expectation is taken over the policy $\pi(h_t)$, and the state transition probability $s_t \sim P$ of POMDP $M$. Our generalization objective in this work is to maximize the discounted cumulative reward taken \textit{in expectation over the POMDP prior}, also termed the \textit{population risk}:
\begin{equation} \label{dist_cumulative_regret}
\mathcal{R}_{pop}(\pi) = \mathbb{E}_{M \sim P(M)}\left[\mathbb{E}_{\pi,M}\left[ \sum_{t=0}^\infty \gamma^tr(s_t,a_t)\right]\right].
\end{equation}
Seeking a policy that performs well in expectation over any POMDP from the prior corresponds to zero-shot generalization.

We assume access to $N$ training POMDPs $M_1,\dots,M_N$ sampled from the prior, $M_i\sim P(M)$. Our goal is to use $M_1,\dots,M_N$ to learn a policy that performs well on objective \ref{dist_cumulative_regret}. 
A common approach is to optimize the \textit{empirical risk} objective:
\begin{equation} \label{ERM}
\mathcal{R}_{emp}(\pi) = \frac{1}{N}\sum_{i=1}^N \mathbb{E}_{\pi,M_i}\left[ \sum_{t=0}^\infty \gamma^tr(s_t,a_t)\right] = \\
\mathbb{E}_{M \sim \hat{P}(M)}\left[\mathbb{E}_{\pi,M}\left[ \sum_{t=0}^\infty \gamma^tr(s_t,a_t)\right]\right],
\end{equation}
where the empirical POMDP distribution can be different from the true distribution, i.e. $\hat{P}(M) \neq P(M)$. In general, a policy that optimizes the empirical risk (Eq.~\ref{ERM}) may perform poorly on the population risk (Eq.~\ref{dist_cumulative_regret})---this is known as overfitting in statistical learning theory~\citep{shalev2014understanding}, and has been analyzed recently also for RL~\citep{tamar2022regularization}.

\subsection*{Maximum Entropy Exploration}

In the following we provide the definitions for the state distribution and the maximum entropy exploration objective. For simplicity, we discuss MDPs---the fully observed special case of POMDPs where $O = S$, and $\Sigma(s,a) = s$.

A policy $\pi$, through its interaction with an MDP, induces a $t$-step state distribution $d_{t,\pi}(s) = p(s_t =s|\pi)$ over the state space $S$. Let $d_{t,\pi}(s,a) = p(s_t =s, a_t=a|\pi)$ be its $t$-step state-action counterpart. For the infinite horizon setting, the stationary state distribution is defined as $d_{\pi}(s) = lim_{t \rightarrow \infty} d_{t,\pi}(s)$, and its $\gamma$-discounted version as $d_{\gamma,\pi}(s) = (1-\gamma)\sum_{t=0}^\infty \gamma^t d_{t,\pi}(s)$. We denote the state marginal distribution as $d_{T,\pi}(s) = \frac{1}{T} \sum_{t=0}^T d_{t,\pi}(s)$, which is a marginalization of the $t$-step state distribution over a finite time $T$. 
The objective of maximum entropy exploration is given by:
\begin{equation} 
\ent(d(\cdot)) = -\mathbb{E}_{s\sim d}[\log(d(s))],
\end{equation} 
where $d$ can be regarded as either the stationary state distribution $d_{\pi}$ \citep{mutti2020intrinsically}, the discounted state distribution $d_{\gamma, \pi}$ \citep{hazan2019provably} or the marginal state distribution $d_{T,\pi}$ \citep{lee2019efficient,mutti2020intrinsically}. In our work we focus on the finite horizon setting and adapt the marginal state distribution $d_{T,\pi}$ in which $T$ equals the episode horizon $H$, i.e. we seek to maximize the objective:
\begin{equation}\label{entropy_onjective}
\mathcal{R}_{\ent}(\pi) = \mathbb{E}_{M \sim \hat{P}(M)}\left[\ent(d_{H,\pi})\right]=\mathbb{E}_{M \sim \hat{P}(M)}\left[\ent\left(\frac{1}{H} \sum_{t=0}^H d_{t,\pi}(s)\right)\right],
\end{equation}
which yields a policy that ``equally'' visits all states during the episode.
Existing works {that target} maximum entropy exploration rely on estimating the density of the agent's state visitation distribution \citep{hazan2019provably,lee2019efficient}. More recently, a branch of algorithms that employ non-parametric entropy estimation \citep{liu2021aps, mutti2021task,seo2021state} has emerged, circumventing the burden of density estimation. 
Here, we follow this common thread and adapt the non-parametric {entropy estimation} approach; we estimate the entropy using the particle-based $k$-nearest neighbor ($k$-NN estimator) \citep{beirlant1997nonparametric,singh2003nearest},
{as elaborated} in the next section. 

\section{The Generalization Ability of Maximum Entropy Exploration}
\label{Generalization_Unsupervised_RL}

In this section we present an empirical observation---policies trained for maximum entropy exploration (maxEnt policy) generalize well. First, we explain the training procedure of our maxEnt policy, then we show empirical results supporting this observation.

\subsection{Training State Space Maximum Entropy Policy}
\label{subsection:training_maxEnt}
To tackle objective (\ref{entropy_onjective}), we estimate the entropy using the particle-based  $k$-NN estimator \citep{beirlant1997nonparametric,singh2003nearest},
{as described here.} Let $X$ be a random variable over the support $\rchi \subset \mathbb{R}^m$ with a probability mass function $p$. Given the probability of this random variable, its entropy is obtained by $\ent_X(p) = -\mathbb{E}_{x \sim p}[\log(p)]$. Without access to its distribution $p$, the entropy can be estimated using $N$ samples $\{x_i\}_{i=1}^N$ by the $k$-NN estimator \cite{singh2003nearest}: 
\begin{equation} \label{sampled_entropy}
\hat{\ent}_X^{k,N}(p)  \approx \frac{1}{N} \sum_{i=1}^N \log\left(\norm{ x_i-x_i^{k\text{-NN}}}_2 \right),
\end{equation} 
where $x_i^{k\text{-NN}}$ is the $k\text{-NN}$ sample of $x_i$ from the set $\{x_i\}_{i=1}^N$.

To estimate the distribution $d_{H,\pi}$ over the states $S$, we consider each trajectory as $H$ samples of states $\{s_t\}_{t=1}^H$ and take $s_t^{k\text{-NN}}$ to be the $k\text{-NN}$ of the state $s_t$ within the trajectory, as proposed by previous works (APT, \cite{liu2021aps}, RE3, \cite{seo2021state}, and APS, \cite{liu2021aps}),
\begin{equation} \label{sampled_entropy_states}
\hat{\ent}^{k,H}(d_{H,\pi}) \approx \frac{1}{H} \sum_{t=1}^H \log\left(	\norm{s_t-s_t^{k\text{-NN}} }_2 \right).
\end{equation} 
Next, similar to previous works, since this sampled estimation of the entropy (Eq.~\ref{sampled_entropy_states}) is a sum of functions that operate on each state separately, it can be considered as an expected reward objective $\hat{\ent}^{k,H}(d_{H,\pi}) \approx \frac{1}{H} \sum_{t=1}^H r_I(s_t)$ with the intrinsic reward function:
\begin{equation} \label{intrinsic_reward}
r_I(s_t) := \log(\norm{s_t-s_t^{k\text{-NN}}}_2).
\end{equation} 
This formulation enables us to deploy any RL algorithm to approximately optimize objective (\ref{entropy_onjective}). Specifically, in our work we use the policy gradient algorithm PPO \citep{schulman2017proximal}, where at every time step $t$ the state $s_t^{k\text{-NN}}$ is chosen from previous states $\{s_i\}_{i=1}^{t-1}$ of the same episode.

Another challenge stems from the computational complexity of calculating the $L_2$ norm of the $k\text{-NN}$ (Eq. \ref{intrinsic_reward}) at every time step $t$.
To improve computational efficiency, we introduce the following approximation: instead of taking the full observation as the state $s_i$ (i.e. $64\times 64$ RGB image), we sub-sample (denoted $\downarrow$) the observation by applying average pooling of $3\times 3$ to produce an image $s_i^{\downarrow}$ of size $21 \times 21$, resulting in:
\begin{equation}\label{intrinsic_reward_approx}
r_I(s_t) := \log\left(\norm{ s_t^{\downarrow}-s_t^{k\text{-NN},\downarrow}}_2\right).
\end{equation} 

\begin{wrapfigure}{r}{0.25\textwidth}
\centering
\includegraphics[width=0.25\textwidth]{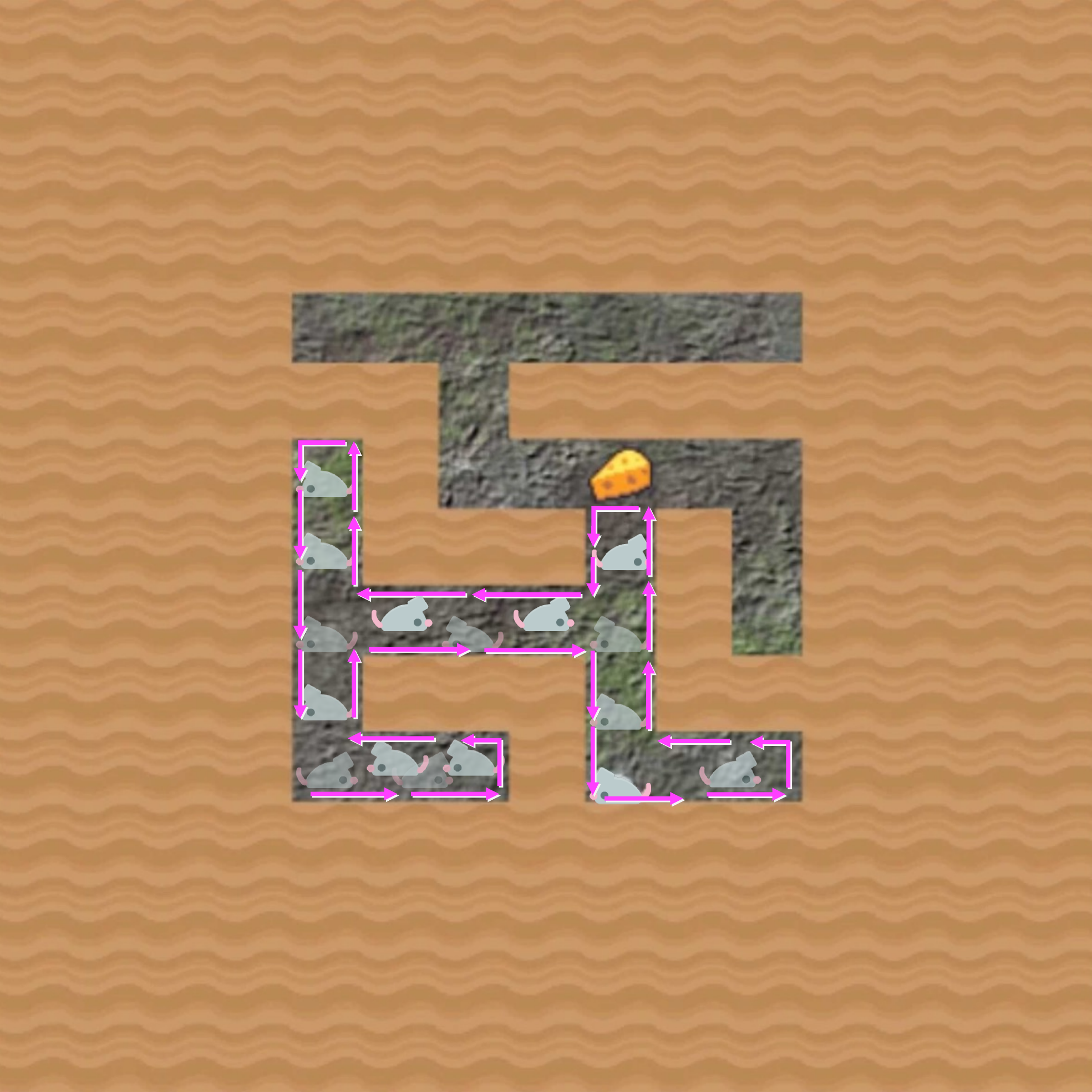}
    \caption{Example of a maxEnt trajectory on Maze. The policy visits every reachable state and averts termination by avoiding the goal state.}
    \label{fig:max_ent_maze}  
    \vspace{-1.1cm}
\end{wrapfigure}

We emphasize that we do not modify the termination condition of each game. However, a maxEnt policy will learn to \textit{avoid} termination, as this increases the sum of intrinsic rewards. In Figure \ref{fig:max_ent_maze} we display the states visited by a maxEnt policy on Maze. 
We also experimented with $L_0$ as the state similarity measure instead of $L_2$, which resulted in similar performance (see Appendix~\ref{appendix:L0_L2_norms}).

\subsection{Generalization of maxEnt Policy}
\label{Sec:gen_gap_maxEnt}

The generalization gap describes the difference between the reward accumulated during training $\mathcal{R}_{emp}(\pi)$ and testing $\mathcal{R}_{pop}(\pi)$ of a policy,
where we approximate the population score by testing on a large population of tasks withheld during training. 
We can evaluate the generalization gap for either an extrinsic reward, or for an intrinsic reward, such as the reward that elicits maxEnt exploration (Eq.~\ref{intrinsic_reward_approx}). In the latter, the generalization gap captures how well the agent's exploration strategy generalizes.

\begin{figure}
\vspace{-1.4cm}
    \centering
     \begin{subfigure}[h]{0.8\textwidth}
        \centering
         \includegraphics[width=0.9\textwidth]{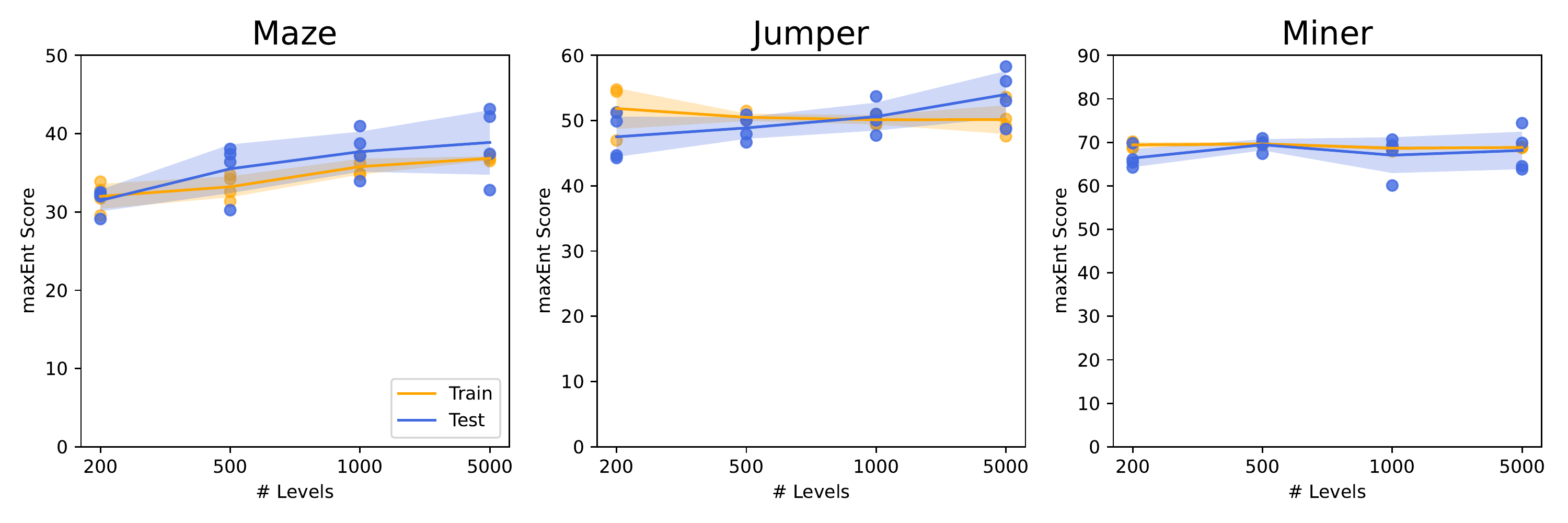}
         \caption{Entropy Reward}
         \label{fig:maxEnt_extReward_entropy}
     \end{subfigure}
     \hfill
     \begin{subfigure}[h]{0.8\textwidth}
        \centering
         \includegraphics[width=0.9\textwidth]
         {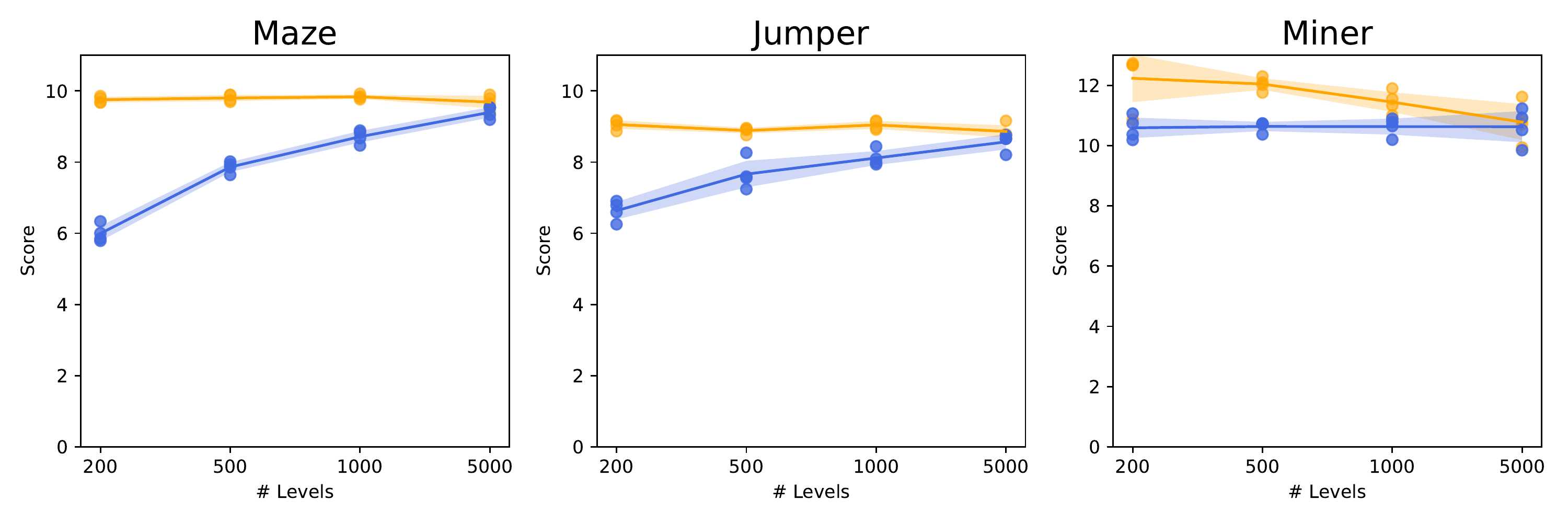}
         \caption{Extrinsic Reward}
         \label{fig:maxEnt_extReward_reward}
     \end{subfigure}
    \caption{\textbf{Generalization ability of maximum entropy vs.~extrinsic reward:} 
    (\textbf{a}) Score of maximum entropy.
    (\textbf{b}) Score of extrinsic reward. 
    Training for maximum entropy exhibits a small generalization gap in Maze, Jumper and Miner. Average and standard deviation are obtained using $4$ seeds.}
    \label{fig:maxEnt_extReward}
    \vspace{-0.4cm}
\end{figure}

We found that agents trained for maximum entropy exploration exhibit a smaller generalization gap compared with the standard approach of training solely with extrinsic reward. Intuitively, this can be attributed to the extrinsic reward serving as an `easy' signal to learn from, and overfit to in the training environments. To assess the generalization quality of the maxEnt policy, we train agents on $200, 500,1000$ and $5000$ instances of ProcGen's Maze, Jumper and Miner environments using the intrinsic reward (Eq.~\ref{intrinsic_reward_approx}). The policies are equipped with a memory unit (GRU,~\citealt{cho2014properties}) to allow learning of deterministic policies that maximize the entropy \citep{mutti2022importance}\footnote{An extensive discussion on the importance of memory for the maxEnt objective is in Appendix \ref{appendix:maxent_memory}.}.

The train and test return scores are shown in Fig.~\ref{fig:maxEnt_extReward_entropy}. In all three environments, we demonstrate a small generalization gap, as test performance on unseen levels closely follows the performance achieved during training. When considering Maze trained on $200$ levels, we observe a small generalization gap of {$1.7\%$}, meaning test performance closely follows train performance. For Jumper and Miner the maxEnt policy exhibits a small generalization gap of {$8.5\%$} and {$4.3\%$}, respectively. In addition, we verify that the train results are near optimal by comparing with a hand designed approximately optimal exploration policy. For example, on Maze we use the well known maze exploring strategy \textit{wall follower}, also known as the left/right-hand rule \citep{hendrawan2020comparison}; see Appendix \ref{appendix:oracel_maxEnt} for details. 

Next, we evaluate the generalization gap of agents trained to maximize the extrinsic reward\footnote{We train for extrinsic reward using an architecture identical to that of the intrinsic reward, with the exception of the memory unit. Incorporating a memory unit in this case further degrades performance (see Appendix \ref{appendix:maxent_memory}).}. The results for this experiment, shown in Fig.~\ref{fig:maxEnt_extReward_reward}, illustrate that the generalization gap for extrinsic reward is more prominent. For comparison, when trained on $200$ levels, the figure shows a large generalization gap for Maze ($38.8\%$) and Jumper ($27.5\%$), while Miner exhibits a moderate generalization gap of $13.1\%$. For an evaluation on all ProcGen games, please see Appendix~\ref{sec:generalization_gap_max_ent_200}.

\section{Explore to Generalize (ExpGen)}
\label{Exploiting_Policy_Uncertainty}

Our main insight is that, given the generalization property of the entropy maximization policy established above, an agent can apply this behavior in a test MDP and expect effective exploration \textit{at test time}. In the following, we pair this insight with the epistemic POMDP idea, and propose to play the exploration policy when the agent faces epistemic uncertainty, hopefully driving the agent to a different state where the reward-seeking policy is more certain. 
This can be seen as an adaptation of the seminal \textit{explicit explore or exploit} idea~\citep{kearns2002near}, to the setting of ZSG-RL.

\subsection{Algorithm}
\label{sec:algorithm}

Our framework comprises two parts: an entropy maximizing network and an ensemble of networks that maximize an extrinsic reward to evaluate epistemic uncertainty.   
\begin{wrapfigure}{R}{0.7\textwidth}
\begin{minipage}{0.7\textwidth}
\vspace{-0.8cm}
\begin{algorithm}[H]
   \caption{\methodname ~(\methodnameshort)}
   \label{algotirhm_1}
\begin{algorithmic}[1]
   \State  {\bfseries Input:} ensemble size $m$,
   \\\qquad\quad
   initial state $s_0 = \textsc{Environment}.\mathrm{reset}()$. 
   \\\qquad\quad $n_{\pi_{\ent}} = 0$
    \State Train maxEnt policy $\pi_{\ent}$ using intrinsic reward $r_I$ (Eq:~\ref{intrinsic_reward}).
   \State Train $m$ policies $\pi_{r}^1,\pi_{r}^2 \ldots\pi_{r}^m$ using extrinsic reward $r_{ext}$.
   \For{$t=1$ {\bfseries to} $H$}
   \State $a_i \sim \pi_{r}^{i}(\cdot|s_t)$
   \State  $n_{\pi_{\ent}} \gets n_{\pi_{\ent}}-1$
   \If{$a_i \in \text{Consensus} (a_j ~|~ j \in \{1 \ldots m\}) ~\mathrm{and}~ n_{\pi_{\ent}} < 0$}  
   \State $a_t = a_i$
   \Else
   \State $a_{\ent}  \sim \pi_{\ent}(\cdot|h_t)$
   \State $a_t = a_{\ent}$
   \State $n_{\pi_{\ent}}\sim {Geom}(\alpha)$
   \EndIf
   \State  $s_{t+1} \gets $~\textsc{Environment.step}$(a_t)$
   \EndFor
\end{algorithmic}
\end{algorithm}
\end{minipage}
\end{wrapfigure}
The first step entails training a network equipped with a memory unit to obtain a maxEnt policy $\pi_{\ent}$ that maximizes entropy, as described in section~\ref{subsection:training_maxEnt}. Next, we train an ensemble of memory-less policy networks $\{\pi_{r}^j\}_{j=1}^m$ to maximize extrinsic reward. Following \citet{ghosh2021generalization}, we shall use the ensemble to assess epistemic uncertainty. Different from \citet{ghosh2021generalization}, however, we do not change the RL loss function, and use an off-the-shelf RL algorithm (such as PPO \citep{schulman2017proximal} or \idaac{}~\citep{stooke2021decoupling}).

At test time, we couple these two components into a combined agent $\boldsymbol{\pi}$ (detailed as pseudo-code in Algorithm~\ref{algotirhm_1}). We consider domains with a finite action space, and say that the policy $\pi_{r}^i$ is certain at state $s$ if its action $a_i\!\sim\! \pi_{r}^i(a|s)$ is in consensus with the ensemble: $a_i=a_j$ for the majority of $k$ out of $m$, where $k$ is a hyperparameter of our algorithm.
When the networks $\{\pi_{r}^j\}_{j=1}^m$ are not in consensus, the agent $\boldsymbol{\pi}$ takes a sequence of $n_{\pi_{\ent}}$ actions from the entropy maximization policy $\pi_{\ent}$, which encourages exploratory behavior. 

\vspace{-0.3cm}
\paragraph{Agent meta-stability} Switching between two policies may result in a case where the agent repeatedly toggles between two states---if, say, the maxEnt policy takes the agent from state $s_1$ to a state $s_2$, where the ensemble agrees on an action that again moves to state $s_1$. To avoid such ``meta-stable'' behavior, we randomly choose the number of maxEnt steps $n_{\pi_{\ent}}$ from a Geometric distribution, $n_{\pi_{\ent}} \sim {Geom}(\alpha)$.

\section{Experiments}
\label{Experiments}
We evaluate our algorithm on the ProcGen benchmark, which employs a discrete 15-dimensional action space and generates RGB observations of size $64 \times 64 \times 3$. Our experimental setup follows ProcGen's `easy' configuration, wherein agents are trained on $200$ levels for $25M$ steps and subsequently tested on random levels~\citep{cobbe2020leveraging}.
All agents are implemented using the IMPALA convolutional architecture \citep{espeholt2018impala}, and trained using PPO~\citep{schulman2017proximal} or \idaac{}~\citep{raileanu2021decoupling}. For the maximum entropy agent $\pi_{\ent}$ we incorporate a single GRU \citep{cho2014properties} at the final embedding of the IMPALA convolutional architecture. For all games, we use the same parameter $\alpha=0.5$ of the Geometric distribution and form an ensemble of $10$ networks.
For further information regarding our experimental setup and specific hyperparameters, please refer to Appendix \ref{appendix:setup}.

% We test our algorithm\footnote{The code will be published with the final version.} on the Procgen benchmark \cite{cobbe2020leveraging} on three environments -- Maze, Heist, and Miner.
% We use the same architecture and hyper-parameters for all experiments as suggested by the authors of the dataset.  

% In our evaluation, we study the performance of \methodnameshort\ and LEEP in terms of accumulated reward over a finite horizon of $H=500$ for the maze environment and $H=1000$ for the heist and miner \textcolor{blue}{ (as suggested by the dataset's authors)}.

\subsection{Generalization Performance}
\label{generalization_performance}
We compare our algorithm to six leading algorithms: vanilla \ppo{}~\citep{schulman2017proximal}, \plr{}~\citep{jiang2021prioritized} that utilizes automatic curriculum-based learning, \ucbdrac{}~\citep{raileanu2021automatic}, which incorporates data augmentation to learn policies invariant to different input transformations,  \ppg{}~\citep{cobbe2021phasic}, which decouples the optimization of policy and value function during learning, and \idaac{}~\citep{raileanu2021decoupling}, the previous state-of-the-art algorithm on ProcGen that decouples policy learning from value function learning and employs adversarial loss to enforce invariance to spurious features. 
Lastly, we evaluate our algorithm against \leep{}~\citep{ghosh2021generalization}, the only algorithm that, to our knowledge, managed to improve upon the performance of vanilla \ppo~on Maze and Heist. The evaluation matches the train and test setting detailed by the contending algorithms and their performance is provided as reported by their authors. For evaluating \leep{} and \idaac{}, we use the original implementation provided by the authors. 
\footnote{For \leep{}~and \idaac{}, we followed the prescribed hyperparameter values of the papers' authors (we directly corresponded with them). For some domains, we could not reproduce the exact results, and in those cases, we used their reported scores, giving them an advantage.}

Tables \ref{table:train_det_nondet25} and \ref{table:test_det_nondet25} show the train and test scores, respectively, for all ProcGen games. The tables show that \methodnameshort~combined with \ppo~achieves a notable gain over the baselines on Maze, Heist and Jumper, while on other games, invariance-based approaches perform better (for example, \idaac{} leads on BigFish, Plunder and Climber, whereas \ppg{}~leads on CaveFlyer, and \ucbdrac{}~leads on Dodgeball). These results correspond to our observation that for some domains, invariance cannot be used to completely resolve epistemic uncertainty. We emphasize that \textit{\methodnameshort\ substantially outperforms \leep{} on all games}, showing that our improved exploration at test time is significant. In Appendix \ref{appendix:table_50} we compare \methodnameshort\ with \leep{} trained for $50M$ environment steps, showing a similar trend. When combining \methodnameshort\ with the leading invariance-based approach, \idaac{}, we establish that \methodnameshort\ is in-fact \textit{complementary} to the advantage of such algorithms, setting a new state-of-the-art performance in ProcGen. A notable exception is Dodgeball, where all current methods still fail.

\begin{figure}[t]
\centering
    \includegraphics[width=0.8\linewidth]{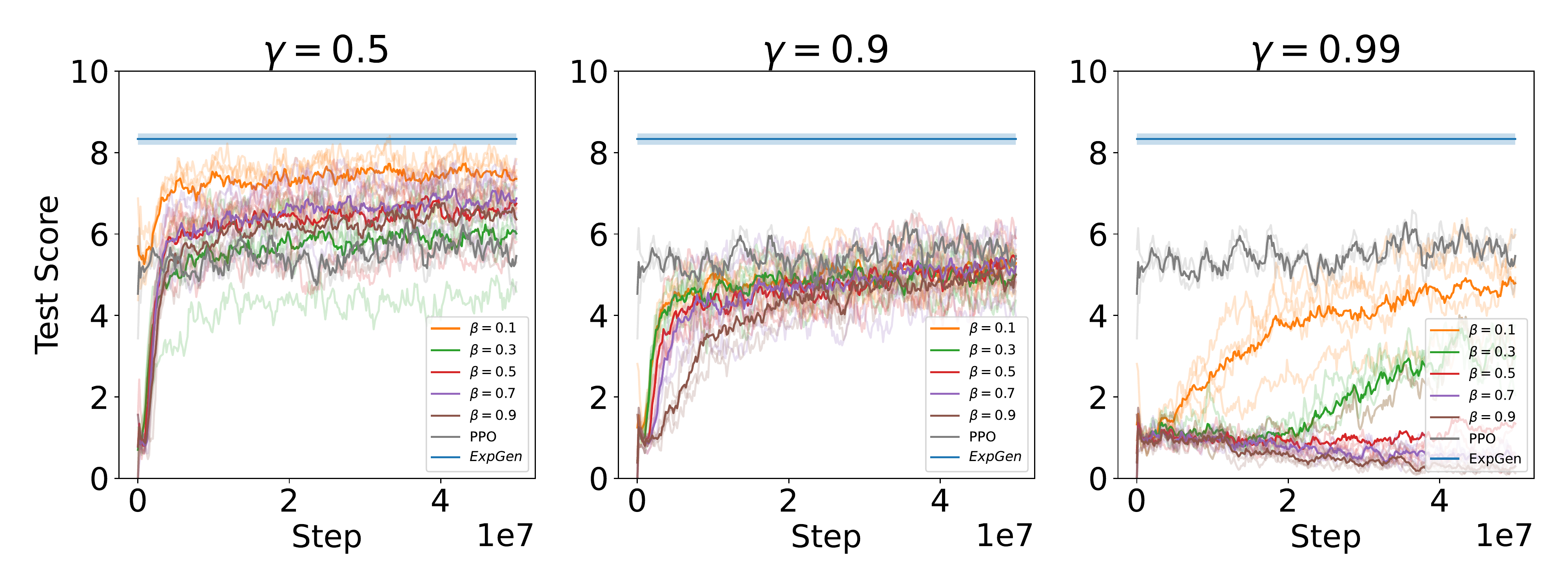}
    \captionsetup{belowskip=-10pt,aboveskip=-2pt}
\caption{Test performance of PPO trained using the reward $r_{total}$ that combines intrinsic and extrinsic rewards, weighted by $\beta$ (Eq.~\ref{eq:r_total}). Each figure details the results for different values of discount factor $\gamma$. All networks are randomly initialized and trained on $200$ maze levels, and their mean is computed over $4$ runs with different seeds. The figures show an improvement over the \ppo{}~baseline for $\gamma=0.5$.
In all cases, \methodnameshort~outperforms the combined reward agent.
} 
\label{fig:ablation_random}
\end{figure}

\begin{figure}[t]
\centering
\begin{subfigure}[h]{0.4\textwidth}
    \centering
     \includegraphics[width=\textwidth]{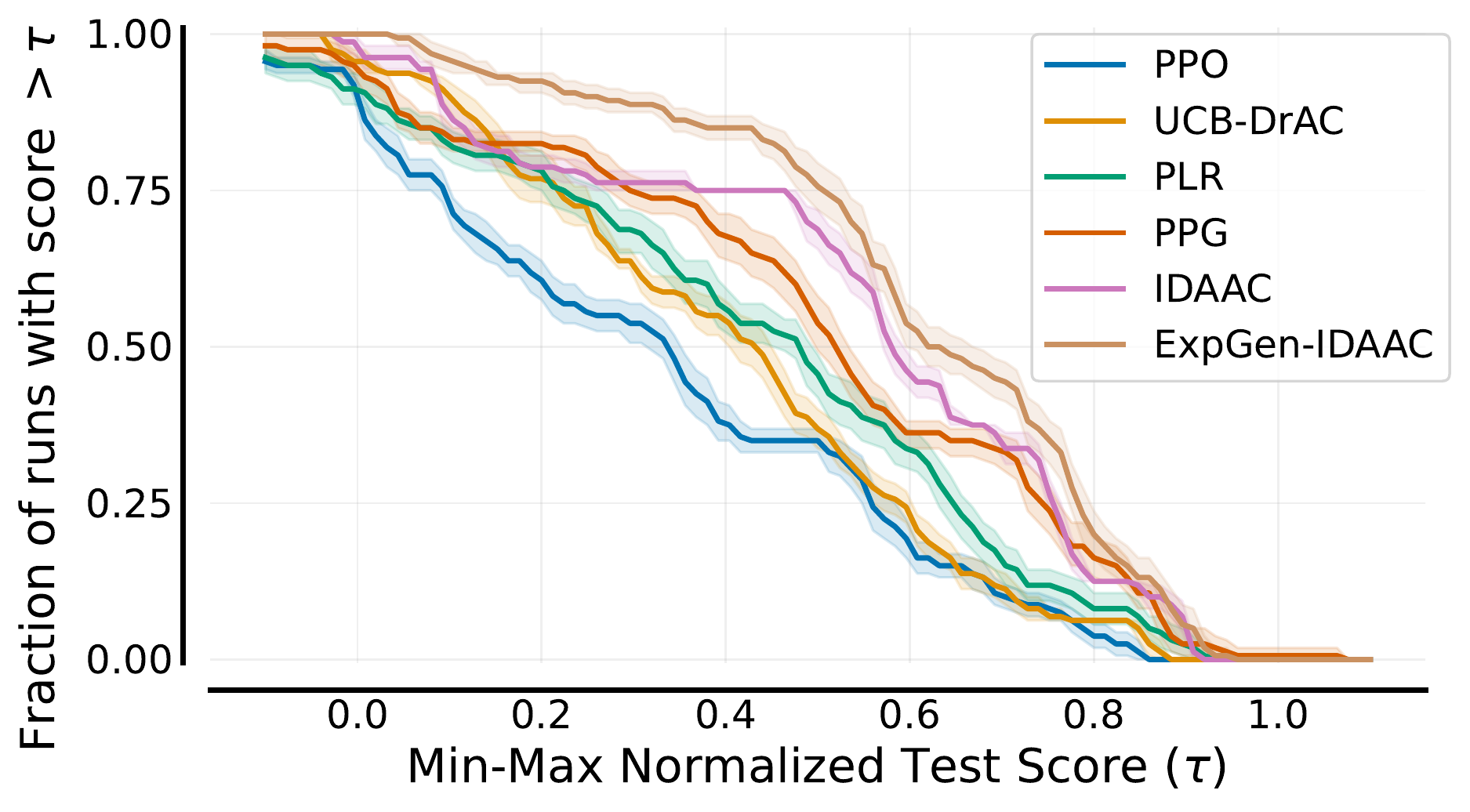}
     \label{procgen_score_distributions}
 \end{subfigure}
 \begin{subfigure}[h]{0.58\textwidth}
    \centering
     \includegraphics[width=\textwidth]{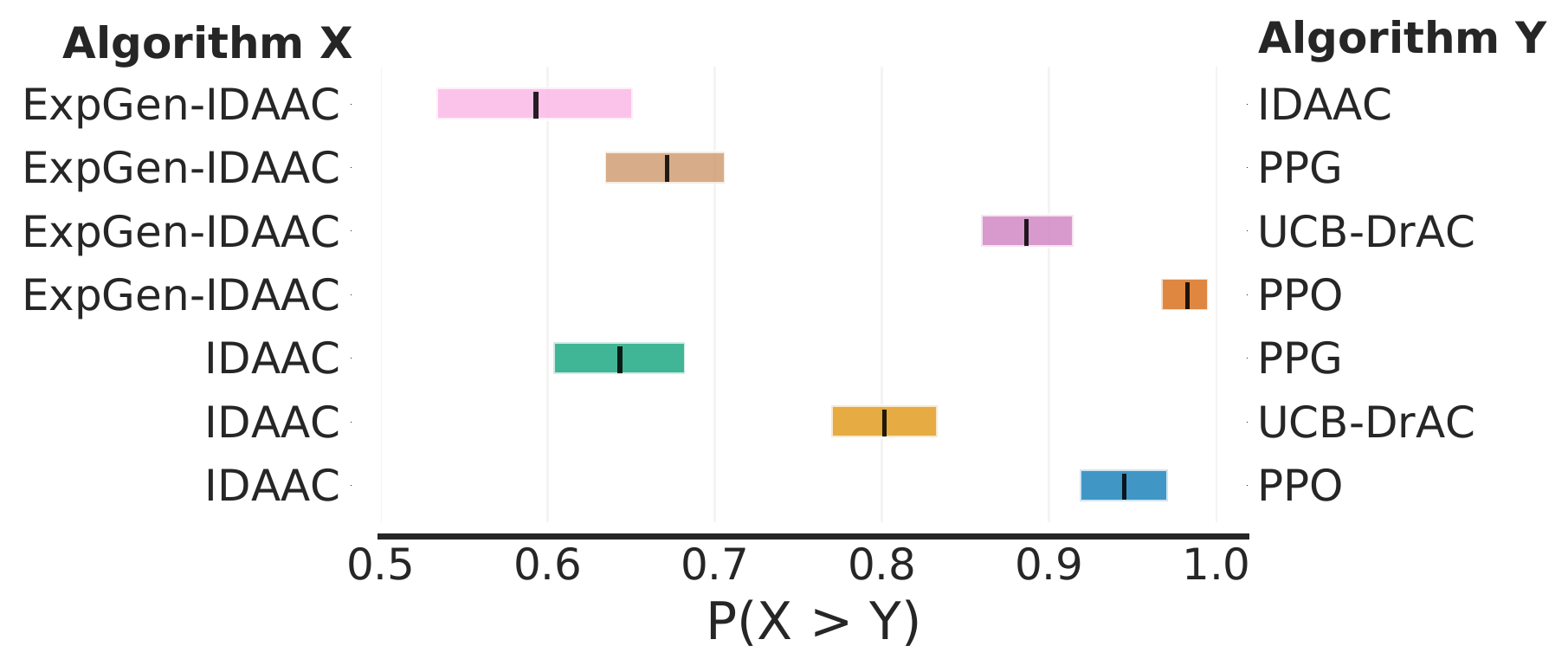}
     \label{procgen_probability}
 \end{subfigure}
 \captionsetup{belowskip=-8pt,aboveskip=-8pt}
\caption{
Comparison across all ProcGen games, with $95\%$ bootstrap CIs highlighted in color. \textbf{Left.} Score distributions of \methodnameshort, \ppo{}, \plr{}, \ucbdrac{},  \ppg{} and \idaac{}. \textbf{Right.} Shows in each row, the probability of algorithm $X$ outperforming algorithm $Y$. The comparison illustrates the superiority of \methodnameshort\ over the leading contender \idaac{} with probability $0.6$, as well as over other methods with even higher probability.
}
\label{fig:probability_of_improvement_score_distributions}
\end{figure}

\begin{figure}[ht]
\centering
    \includegraphics[width=1.01\linewidth]{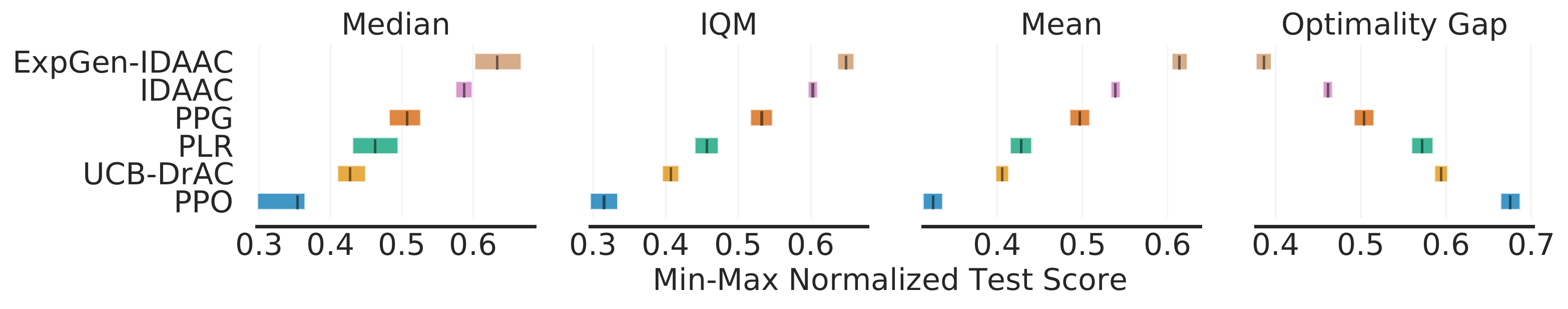}
    \captionsetup{belowskip=-8pt,aboveskip=-2pt}
\caption{
Aggregate metrics for all ProcGen games: mean, median and IQM scores (higher is better) and optimality gap (lower is better), with $95\%$ CIs highlighted in color. \methodnameshort\ outperforms the contending methods in all measures.
} 
\label{fig:aggregate_metrics}
\end{figure}

Figures~\ref{fig:probability_of_improvement_score_distributions} and \ref{fig:aggregate_metrics} show aggregate statistics of \methodnameshort, \ppo{}, \plr{}, \ucbdrac{},  \ppg{} and \idaac{} for all games
\footnote{See \textit{rliable} \citep{agarwal2021deep} for additional details on the various performance measures and protocols.}, affirming the dominance of \methodnameshort+\idaac{} as the state-of-the-art. The results are obtained using $10$ runs per game, with scores normalized as in Appendix~\ref{appendix:norm_constants}. The shaded regions indicate $95\%$ Confidence Intervals (CIs) and are estimated using the percentile stratified bootstrap with $2,000$ (Fig.~\ref{fig:probability_of_improvement_score_distributions}) and $50,000$ (Fig.~\ref{fig:aggregate_metrics}) bootstrap re-samples. 
Fig.~\ref{fig:probability_of_improvement_score_distributions} (Left) compares algorithm score-distribution, illustrating the advantage of the proposed approach across all games.
Fig.~\ref{fig:probability_of_improvement_score_distributions} (Right) shows the probability of improvement of algorithm $X$ against algorithm $Y$. The first row (\methodnameshort\ vs.~\idaac{}) demonstrates that the proposed approach surpasses \idaac{} with probability $0.6$ and subsequent rows emphasize the superiority of \methodnameshort\ over contending methods at an even higher probability. This is because \methodnameshort~improves upon \idaac{} in several key challenging tasks and is on-par in the rest. Fig.~\ref{fig:aggregate_metrics} provides aggregate metrics of mean, median and IQM scores and optimality gap (as $1-mean$) for all ProcGen games. The figure shows that \methodnameshort\ outperforms the contending methods in all measures.

\paragraph{Ablation Study}
\label{ablation_study}
One may wonder if the ensemble in \methodnameshort\ is necessary, or whether the observation that the maxEnt policy generalizes well can be exploited using a single policy.
We investigate the effect of combining the intrinsic and extrinsic rewards, $r_I$ and $r_{ext}$, respectively, into a single reward as a weighted sum:
\begin{equation}\label{eq:r_total}
    r_{\mathrm{total}}=\beta r_I + (1 - \beta) r_{\mathrm{ext}},
\end{equation}
and train for $\beta=\{0.1, 0.3, 0.5, 0.7, 0.9\}$ on Maze.
Figure \ref{fig:ablation_random} shows the train and test scores over $50M$ steps for different values of discount factor $\gamma$. We obtain the best test score for $\gamma=0.5$ and $\beta=0.1$, illustrating an improvement compared with the PPO baseline. When comparing with \methodnameshort, the combined reward (Eq. \ref{eq:r_total}) exhibits inferior performance with slightly higher variance.
In Appendix \ref{appendix:hyperparameters} and \ref{appendix:ablation_study}, we also provide an ablation study of ensemble size and draw comparisons to variants of our algorithm.

\begin{table*}[ht]
    \centering
    \resizebox{\linewidth}{!}{%
    \begin{tabular}{l|cccccc|cc}
    \toprule
    Game & PPO & \plr{}	& \ucbdrac{}  & \ppg{}  & \idaac{} & \leep{} & \methodnameshort & \methodnameshort\\
     &   & 	&   &   &  &  & (PPO) & (IDAAC)\\
    \toprule
    BigFish & $2.9\pm1.1 $   & $10.9\pm2.8$  &	$9.2\pm2.0$ &  $11.2\pm1.4$ &	$\pmb{18.5\pm1.2}$ & $4.9\pm0.9$& ${6.0\pm0.5}$ & ${\pmb{18.5\pm1.9}}$   \\
    StarPilot& $24.9\pm1.0 $  & $27.9\pm4.4$ &	$30.0\pm1.3$ & $\pmb{47.2\pm1.6}$ &	$37.0\pm2.3$ &$3.2\pm2.2$ & ${31.0\pm0.9}$	& ${39.8\pm2.9}$	  \\
    FruitBot& $26.2\pm1.2 $   & $28.0\pm1.4$  &$27.6\pm0.4$ & $27.8\pm0.6$ & 	$\pmb{27.9\pm0.5}$ & $16.4\pm1.6$	& ${26.2\pm0.6}$ & ${\pmb{28.4\pm0.4}}$
  \\
    BossFight& $7.4\pm0.4$   & $8.9\pm0.4$  &$7.8\pm0.6 $ & $\pmb{10.3\pm0.2}$ & 	$9.8\pm0.6$ & $0.5\pm0.3$	&${7.7\pm0.2}$ &${9.8\pm0.5}$
  \\
    Ninja&      $6.1\pm0.2$   & $\pmb{7.2\pm0.4}$ &	$6.6\pm0.4 $ &  $6.6\pm0.1$	 & $6.8\pm0.4$	 & $4.4\pm0.5$ &${6.6\pm0.2}$ &${6.6\pm0.3}$
  \\
    Plunder&    $7.8\pm1.6$  &  $8.7\pm2.2$ &	$8.3\pm1.1$ &  $14.3\pm2.0$ & 	$\pmb{23.3\pm1.4}$ &  $4.4\pm0.3$& ${5.5\pm1.3}$	& ${\pmb{23.6\pm1.4}}$
  \\
    CaveFlyer& $5.5\pm0.5$ & $6.3\pm0.5$  &		$5.0\pm0.8$  & $\pmb{7.0\pm0.4}$	 & $5.0\pm0.6$ & 	$4.9\pm0.2$ &${5.7\pm0.3}$&${5.3\pm0.7}$
  \\
    CoinRun&	$8.6\pm0.2 $   &  $8.8\pm0.5$  &$8.6\pm0.2$ & $8.9\pm0.1$ & 	$\pmb{9.4\pm0.1}$ & $7.3\pm0.4$ &${8.8\pm0.1}$ &$\pmb{9.3\pm0.3}$
  \\
    Jumper&	  $5.8\pm0.3 $  &  $5.8\pm0.5$ &	$6.2\pm0.3$&  $5.9\pm0.1$	 & $6.3\pm0.2$ & 	$5.4\pm1.2$ &${\pmb{6.7\pm0.3}}$ &${\pmb{6.8\pm0.5}}$
 \\
    Chaser&	  $3.1\pm0.9 $  & $6.9\pm1.2$  &	$6.3\pm0.6$&  $\pmb{9.8\pm0.5}$	 & $6.8\pm1.0$ & $3.0\pm0.1$ &${3.6\pm1.6}$ &${7.1\pm1.4}$	
  \\
    Climber&	 $5.4\pm0.5 $   & $6.3\pm0.8$  & $6.3\pm0.6 $  & $2.8\pm0.4$ & 	${8.3\pm0.4}$	 & $2.6\pm0.9$ & ${5.9\pm0.5}$ & ${\pmb{9.5\pm0.3}}$
 \\
    Dodgeball& $2.2\pm0.4 $   &  $1.8\pm0.5$  &	$\pmb{4.2\pm0.9}$& $2.3\pm0.3$ & 	$3.2\pm0.3$ & 	$1.9\pm0.2$ & ${2.9\pm0.3}$ & ${2.8\pm0.2}$
 \\
    Heist&	    $2.4\pm0.5 $   & $2.9\pm0.5$  &	$3.5\pm0.4$ &  $2.8\pm0.4$	 & $3.5\pm0.2$ &  $4.5\pm0.3$	& ${\pmb{7.4\pm0.2}}$ & ${{\pmb{7.2\pm0.5}}}$
  \\
    Leaper&	    $4.9\pm2.2 $  &  $6.8\pm1.2$  &	$4.8\pm0.9$ & $\pmb{8.5\pm1.0}$	 & $7.7\pm1.0$ & $4.4\pm0.2$	&${4.0\pm1.7}$ &${7.6\pm1.2}$
 \\
    Maze&	     $5.6\pm0.1$  &  $5.5\pm0.8$ &	$6.3\pm0.1$ & $5.1\pm0.3$ & 	$5.6\pm0.3$	 &  $6.6\pm0.2$ & ${\pmb{8.3\pm0.2}}$ & ${{7.8\pm0.2}}$
 \\
    Miner&	 $7.8\pm0.3 $  &  $9.6\pm0.6$ &		$9.2\pm0.6$ &  $7.4\pm0.2$ & 	$\pmb{9.5\pm0.4}$ & 	$1.1\pm0.1$ & ${8.0\pm0.7}$ & ${\pmb{9.8\pm0.3}}$
  \\
    \bottomrule
    \end{tabular}
    } % resizebox
    \caption{\textbf{Test score} of ProcGen games trained on $200$ levels for $25M$ environment steps. We compare our algorithm to \ppo{}, \plr{}, \ucbdrac{},  \ppg{}, \idaac{} and \leep{}. The mean and standard deviation are computed over $10$ runs with different seeds.}
    \label{table:test_det_nondet25}
\end{table*}

\begin{table*}[ht]
    \centering
    \resizebox{\linewidth}{!}{%
    \begin{tabular}{l|cccccc|cc}
    \toprule
    Game & PPO & \plr{}	& \ucbdrac{}  & \ppg{}  & \idaac{} & \leep{} & \methodnameshort & \methodnameshort\\
     &   & 	&   &   &  &  & (PPO) & (IDAAC)\\
    \toprule
    BigFish & $8.9\pm2.0 $  & $7.8\pm1.0$ &	$12.8\pm1.8$ & $19.9\pm1.7$ & 	$\pmb{21.8\pm1.8}$ & $8.9\pm0.9$ & ${7.0\pm0.4}$	& $\pmb{21.5\pm2.3}$
 \\
    StarPilot& $29.0\pm1.1 $   & $2.6\pm0.3$ &		$33.1\pm1.3$ & $\pmb{49.6\pm2.1}$ & 	$38.6\pm2.2$ & $5.3\pm0.3$	&${34.3\pm1.6}$ &${40.0\pm2.7}$
  \\
    FruitBot& $28.8\pm0.6 $   &  $15.9\pm1.3$ &	$29.3\pm0.5$ &  $\pmb{31.1\pm0.5}$ & 	$29.1\pm0.7$ & $17.4\pm0.7$	&${28.9\pm0.6}$ &${29.5\pm0.5}$
  \\
    BossFight& $8.0\pm0.4 $  & $8.7\pm0.7$ &	$8.1\pm0.4$ & $\pmb{11.1\pm0.1}$ & 	$10.4\pm0.4$ & $0.3\pm0.1$	&${7.9\pm0.6}$ &${9.9\pm0.7}$
 \\
    Ninja&      $7.3\pm0.2 $   & $5.4\pm0.5$ &	$8.0\pm0.4$ &  $8.9\pm0.2$ & 	$\pmb{8.9\pm0.3}$ & $4.6\pm0.2$ & ${8.5\pm0.3}$	& ${7.9\pm0.6}$
  \\
    Plunder&    $9.4\pm1.7 $   &  $4.1\pm1.3$ &	$10.2\pm1.8$  & $16.4\pm1.9$ & 	$\pmb{24.6\pm1.6}$ & $4.9\pm0.2$ & ${5.8\pm1.4}$ & ${\pmb{26.1\pm2.7}}$
  \\
    CaveFlyer& $7.3\pm0.7 $  & $6.4\pm0.1$ &		$5.8\pm0.9$  & $\pmb{9.5\pm0.2}$ & 	$6.2\pm0.6$ & 	$4.9\pm0.3$ & ${6.8\pm0.4}$ & ${5.5\pm0.5}$
 \\
    CoinRun&	$9.4\pm0.3 $  & $5.4\pm0.4$ &		$9.4\pm0.2$ & $\pmb{9.9\pm0.0}$ & 	$9.8\pm0.1$ & $6.7\pm0.1$	& ${9.8\pm0.1}$ & ${9.1\pm0.4}$
 \\
    Jumper&	  $8.6\pm0.1 $  &  $3.6\pm0.5$ &		$8.2\pm0.1$  & $8.7\pm0.1$ & $\pmb{8.7\pm0.2}$ & $5.7\pm0.1$ & ${7.9\pm0.2}$ & ${8.1\pm0.4}$
  \\
    Chaser&	  $3.7\pm1.2 $ & $6.3\pm0.7$ &		$7.0\pm0.6$  & $\pmb{10.7\pm0.4}$	 & $7.5\pm0.8$ & $2.6\pm0.1$	&${4.7\pm1.8}$ &${6.9\pm1.1}$
  \\
    Climber&	 $6.9\pm1.0 $   &  $6.2\pm0.8$ &		$8.6\pm0.6$ & $10.2\pm0.2$ & ${10.2\pm0.7}$ & $3.5\pm0.3$ & ${7.7\pm0.4}$ & ${\pmb{11.4\pm0.2}}$
 \\
    Dodgeball& $6.4\pm0.6$ &  $2.0\pm1.1$ &		$\pmb{7.3\pm0.8}$ &  $5.5\pm0.5$ & 	$4.9\pm0.3$ & 	$3.3\pm0.1$ & ${5.8\pm0.5}$ & ${5.3\pm0.5}$
  \\
    Heist&	    $6.1\pm0.8$ & $1.2\pm0.4$ &		$6.2\pm0.6$ & $7.4\pm0.4$ & 	$4.5\pm0.3$	 & $7.1\pm0.2$ & ${\pmb{9.4\pm0.1}}$ & ${7.0\pm0.6}$
  \\
    Leaper&	    $5.5\pm0.4$  & $6.4\pm0.4$  &		$5.0\pm0.9$ & $\pmb{9.3\pm1.1}$ & 	$8.3\pm0.7$ & $4.3\pm2.3$	&${4.3\pm2.0}$ &${8.1\pm0.9}$
  \\
    Maze&	     $9.1\pm0.2 $  & $4.1\pm0.5$ &		$8.5\pm0.3$ &  $9.0\pm0.2$ & 	$6.4\pm0.5$ & 	$9.4\pm0.3$ & ${\pmb{9.6\pm0.1}}$ & ${{7.4\pm0.4}}$
  \\
    Miner&	 $11.3\pm0.3 $  & $9.7\pm0.4$ &	$\pmb{12.0\pm0.3}$ &  $11.3\pm1.0$ & 	$11.5\pm0.5$ & 	$1.9\pm0.6$ & ${9.0\pm0.8}$& ${11.9\pm0.2}$
 \\
    \bottomrule
    \end{tabular}
    } % resizebox
    \caption{\textbf{Train score} of ProcGen games trained on $200$ levels for $25M$ environment steps. We compare our algorithm to \ppo{}, \plr{}, \ucbdrac{},  \ppg{}, \idaac{} and \leep{}. The mean and standard deviation are computed over $10$ runs with different seeds.}
    \label{table:train_det_nondet25}
\end{table*}

\section{Discussion and Limitations}
We observed that policies trained to explore, using maximum entropy RL, exhibited generalization of their exploration behavior in the zero-shot RL setting. Based on this insight, we proposed \methodnameshort ---a ZSG-RL algorithm that takes a maxEnt exploration step whenever an ensemble of policies trained for reward maximization does not agree on the current action. We demonstrated that this simple approach performs well on {all} ZSG-RL domains of the ProcGen benchmark.

One burning question is \textit{why does maxEnt exploration generalize so well?} An intuitive argument is that the maxEnt policy in an MDP is \textit{invariant} to the reward. Thus, if for every training MDP there are many different rewards, each prescribing a different behavior, the maxEnt policy has to be invariant to this variability. In other words, the maxEnt policy contains \textit{no information} about the rewards in the data, and generalization is well known to be bounded by the mutual information between the policy and the training data~\citep{bassily2018learners}. Perhaps an even more interesting question is whether the maxEnt policy is also less sensitive to variations in the dynamics of the MDPs. We leave this as an open theoretical problem.

Another consideration is safety. In some domains, a wrong action can lead to a disaster, and in such cases, exploration at test time should be hedged. One possibility is to add to \methodnameshort's policy ensemble an ensemble of advantage functions, and use it to weigh the action agreement~\citep{rotman2020online}. Intuitively, the ensemble should agree that unsafe actions have a low advantage, and not select them at test time.

Finally, we point out that while our work made significant progress on generalization in several ProcGen games, the performance on Dodgeball remains low for all methods we are aware of. An interesting question is whether performance on Dodgeball can be improved by combining invariance-based techniques (other than \idaac{}) with exploration at test time, or whether Dodgeball represents a different class of problems that requires a completely different approach.

% At present, our results for maxEnt generalization are purely empirical. Whether such generalization can be theoretically motivated is an interesting direction for future research. Additionally, an intriguing direction for further investigation involves integrating \methodnameshort~with concepts inspired by algorithms that excel in different games (e.g., BigFish, Plunder, etc.) to create a unified agent that consistently outperforms across all games.

\vspace{0.6cm}
\paragraph{Acknowledgments}
The research of DS was Funded by the European Union (ERC, A-B-C-Deep, 101039436). 
The research of EZ and AT was Funded by the European Union (ERC, Bayes-RL, 101041250). 
Views and opinions expressed are however those of the author(s) only and do not necessarily reflect those of the European Union or the European Research Council Executive Agency (ERCEA). Neither the European Union nor the granting authority can be held responsible for them. DS also acknowledges the support of the Schmidt Career Advancement Chair in AI.

\medskip

\bibliographystyle{plainnat}
\bibliography{expgen_bib.bib}

% {
% \small

% [1] Alexander, J.A.\ \& Mozer, M.C.\ (1995) Template-based algorithms for
% connectionist rule extraction. In G.\ Tesauro, D.S.\ Touretzky and T.K.\ Leen
% (eds.), {\it Advances in Neural Information Processing Systems 7},
% pp.\ 609--616. Cambridge, MA: MIT Press.

% [2] Bower, J.M.\ \& Beeman, D.\ (1995) {\it The Book of GENESIS: Exploring
%   Realistic Neural Models with the GEneral NEural SImulation System.}  New York:
% TELOS/Springer--Verlag.

% [3] Hasselmo, M.E., Schnell, E.\ \& Barkai, E.\ (1995) Dynamics of learning and
% recall at excitatory recurrent synapses and cholinergic modulation in rat
% hippocampal region CA3. {\it Journal of Neuroscience} {\bf 15}(7):5249-5262.
% }

%%%%%%%%%%%%%%%%%%%%%%%%%%%%%%%%%%%%%%%%%%%%%%%%%%%%%%%%%%%%
\appendix

\clearpage

\section{Hidden Maze Experiment}\label{appendix:hidden_maze_exp}

\begin{figure}[h]
    \centering
     \begin{subfigure}[h]{\textwidth}
        \centering
         \includegraphics[width=\textwidth]{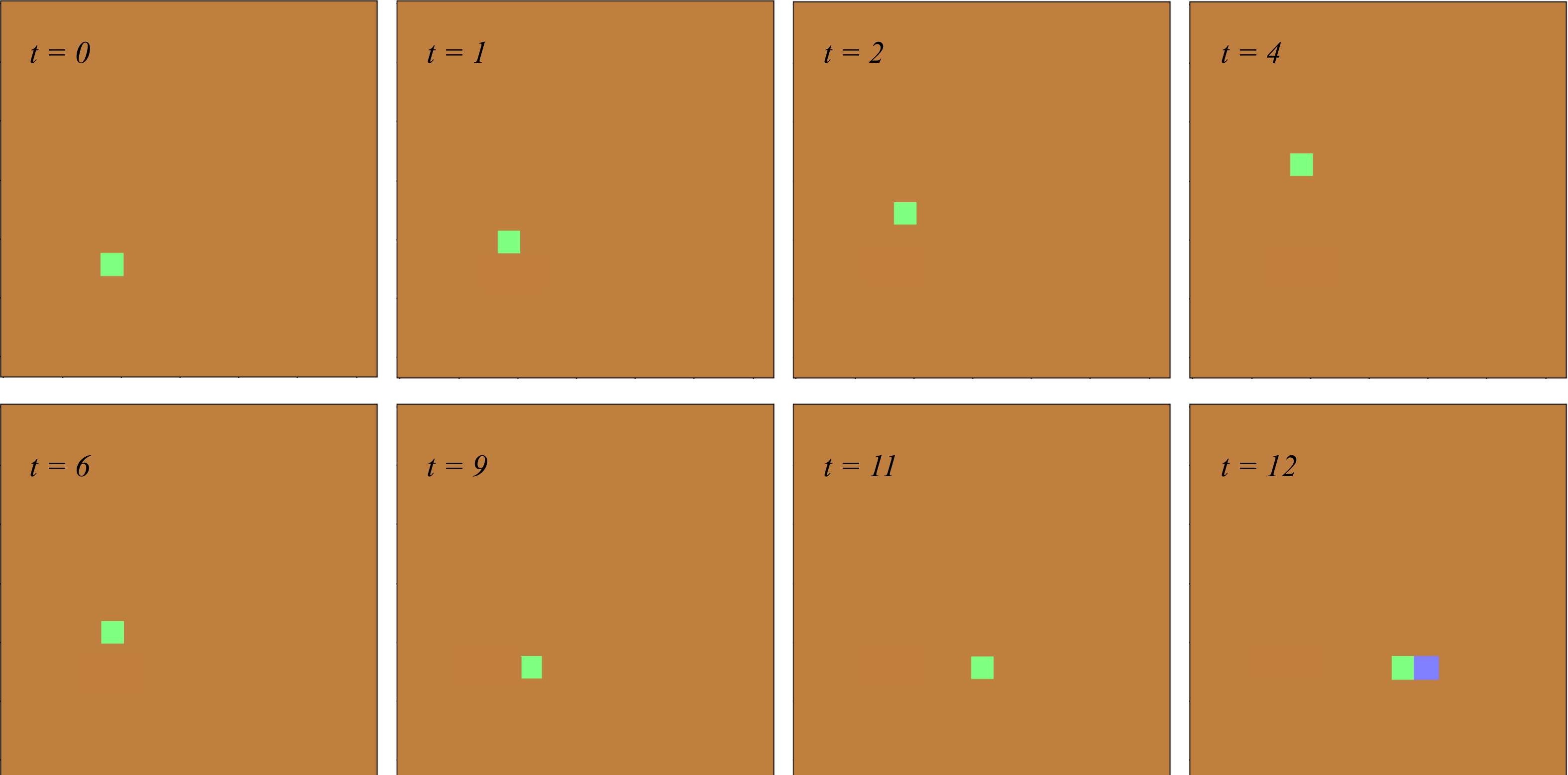}
          \caption{Agent observations along a trajectory for the Hidden Maze task. The bottom-right frame ($t=12$) shows the agent eventually revealing the goal.}
         \label{fig:hidden_maze_obs}
     \end{subfigure}
     \begin{subfigure}[h]{\textwidth}
        \centering
         \includegraphics[width=\textwidth]{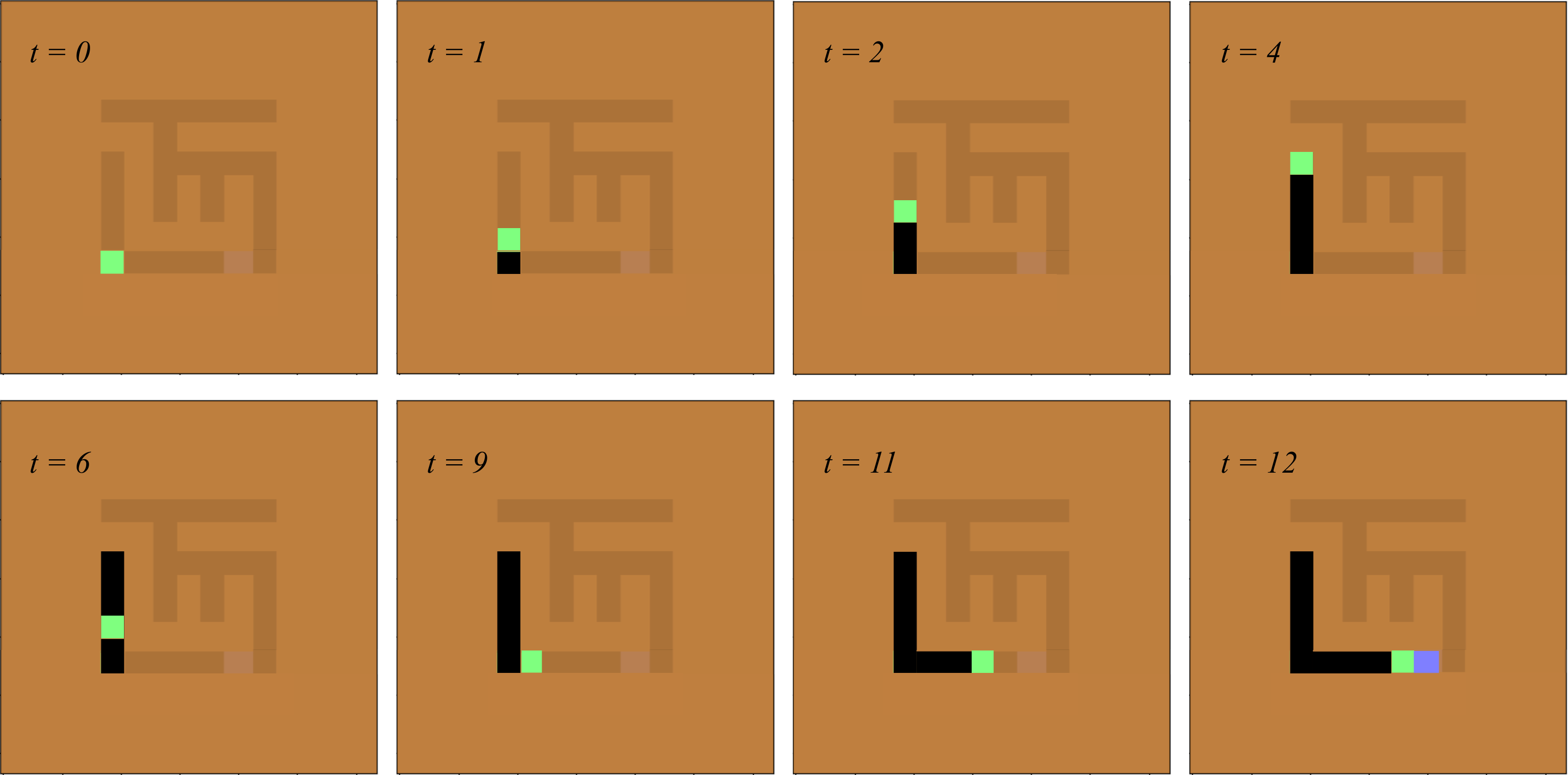}
          \caption{The corresponding \textit{unobserved} true states of agent. Unexplored regions of the maze appear as faded sections.}
         \label{fig:hidden_maze_true_state}
     \end{subfigure}
    \caption{Hidden Maze experiment where the agent only observes its own location (green spot). Both the goal (purple spot) and corridors are not observable (appear as walls).}
    \label{fig:hidden_maze}
\end{figure}

\begin{figure}[ht]
\centering
    \includegraphics[width=0.5\linewidth]{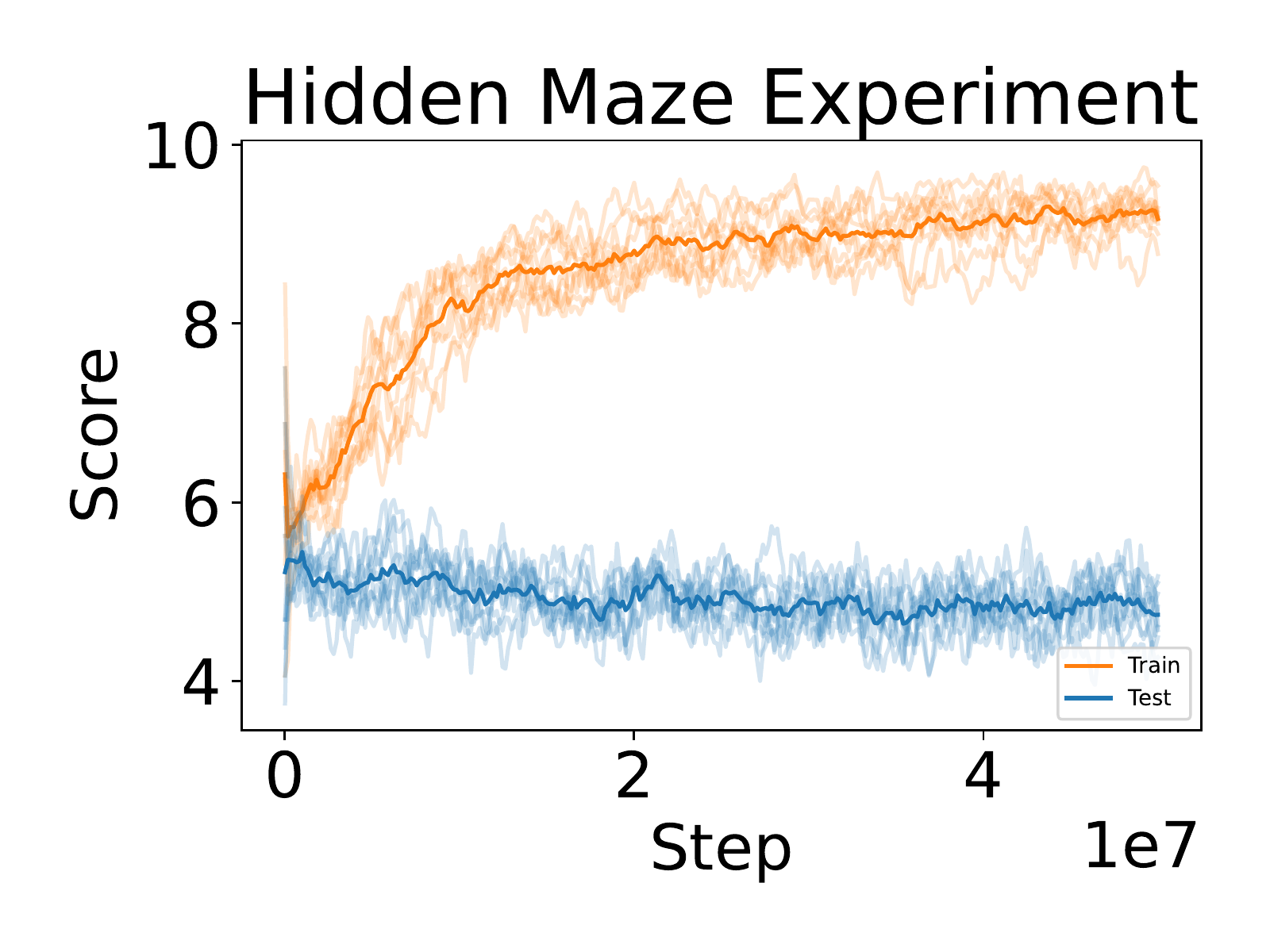}
\caption{\ppo{} performance on the hidden maze task, indicating severe overfitting. Train (Orange) and Test (blue) performance are displayed for $10$ seeds, alongside their means (highlighted in bold).
} 
\label{fig:hidden_maze_graph}
\end{figure}

In this section we empirically validate our thought experiment described in the Introduction: we train a recurrent policy on hidden mazes using 128 training levels (with fixed environment colors).
Figure~\ref{fig:hidden_maze_obs} depicts the observations of an agent along its trajectory; the agent sees only its own location (green spot), whereas the entire maze layout, corridors and goal, are hidden. In Fig.~\ref{fig:hidden_maze_true_state} we visualize the full, unobserved, state of the agent. The train and test results are shown in Fig~\ref{fig:hidden_maze_graph}, indicating severe overfitting to the training levels: test performance failed to improve beyond the random initial policy during training. Indeed, the recurrent policy memorizes the agent's training trajectories instead of learning generalized behavior. Hiding the maze's goal and corridors leads to agent-behavior that is the most invariant to instance specific observations---the observations include the minimum information, such that the agent can still learn to solve the maze. 

This experiment demonstrates why methods based on observation invariance (e.g., \idaac{}~\cite{raileanu2021decoupling}) do not improve performance on Maze-like tasks, despite significantly improving performance on games such as BigFish and Plunder, where invariance to colors helps to generalize.

\section{Maximum Entropy Policy}
This section elaborates on the implementation details of the maxEnt oracle and provides a performance evaluation of the maxEnt policy.

\subsection{Generalization Gap of maxEnt vs PPO across all ProcGen Environments}
\label{sec:generalization_gap_max_ent_200}
\begin{figure}[h!]
\centering
\includegraphics[width=.8\textwidth]{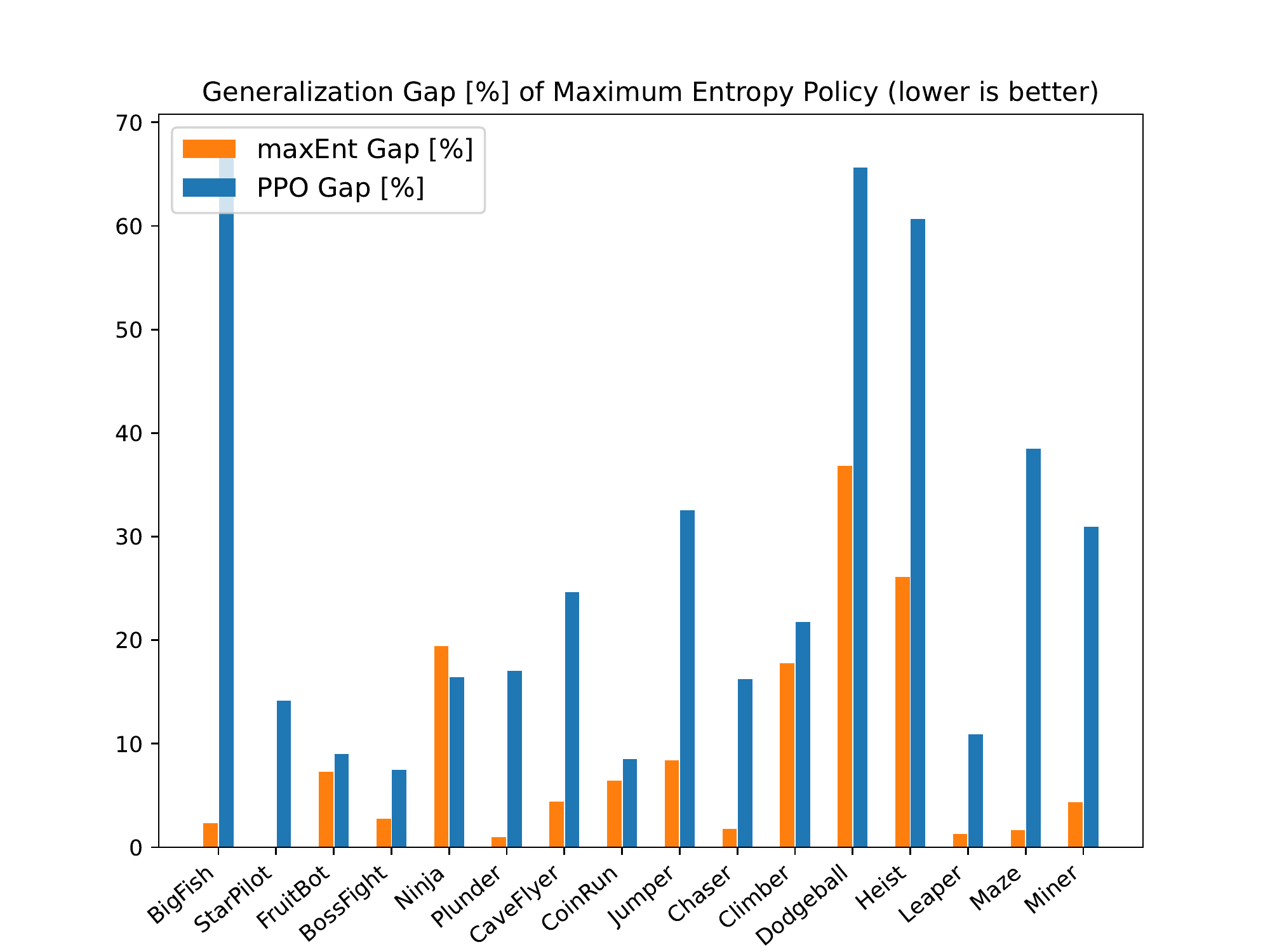}
\caption{The normalized generalization gap [\%] of maxEnt and \ppo{}~for all ProcGen games, trained on $200$ training levels and averaged over $4$ seeds (lower is better).}
\label{fig:gen_gap_maxent_ppo_200}
\end{figure}

Recall from Section \ref{Sec:gen_gap_maxEnt} that the (normalized) generalization gap is described by $$(\mathcal{R}_{emp}(\pi)-\mathcal{R}_{pop}(\pi))/\mathcal{R}_{emp}(\pi),$$ where ${R}_{emp}(\pi)=\mathcal{R}_{train}(\pi)$ and ${R}_{pop}(\pi)=\mathcal{R}_{test}(\pi)$. Fig~\ref{fig:gen_gap_maxent_ppo_200} shows the generalization ability of the maxEnt exploration policy compared to \ppo{}, obtained from training on $200$ training levels.

{The figure demonstrates that the maxEnt exploration policies transfer better in zero-shot generalization (achieve a smaller generalization gap) across all ProcGen games apart from Ninja. This holds true even in environments where the ExpGen algorithm is on par but does not exceed the baseline, pointing to the importance of exploratory behavior in some environments, but not in others.}

\subsection{Computing the maxEnt Oracle}
\label{appendix:oracel_maxEnt}
The maxEnt oracle prescribes the maximum intrinsic return achievable per environment instance.
To simplify the computation of the maxEnt oracle score, in this section, we evaluate the maxEnt score using the $L_0$ instead of the $L_2$ norm and use the first nearest neighbor ($k=1$).
In Maze, we implement the oracle using the right-hand rule~\citep{hendrawan2020comparison} for maze exploration: If there is no wall and no goal on the right, the agent turns right, otherwise, it continues straight. If there is a wall (or goal) ahead, it goes left (see Figure~\ref{fig:max_ent_maze}). In Jumper, we first down-sample (by average pooling) from $64\times 64$ to $21\times 21$ pixels to form cells of $3 \times 3$, and the oracle score is the number of background cells (cells that the agent can visit). Here we assume that the agent has a size of $3 \times 3$ pixels.
In Miner, the ``easy'' environment is partitioned into $10\times 10$ cells. The maximum entropy is the number of cells that contain ``dirt'' (i.e., cells that the agent can excavate in order to make traversable).

\begin{figure}[ht]
    \centering
     \begin{subfigure}[h]{1.\textwidth}
        \centering
         \includegraphics[width=0.9\textwidth]{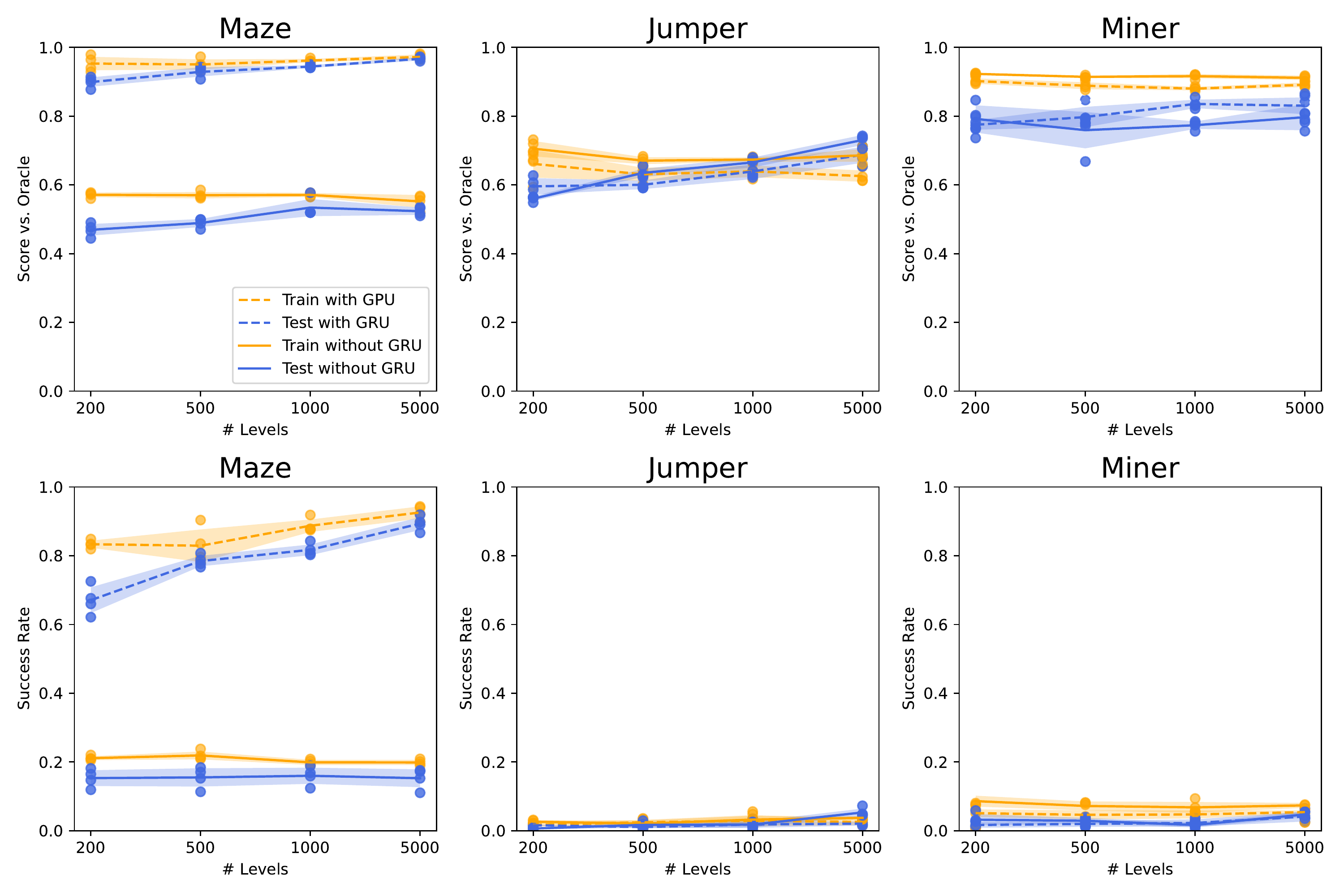}
         \caption{Entropy Reward}
         \label{fig:maxEnt_extReward_entropy_withRNN}
     \end{subfigure}
     \hfill
     \begin{subfigure}[h]{1.\textwidth}
        \centering
         \includegraphics[width=0.9\textwidth]
         {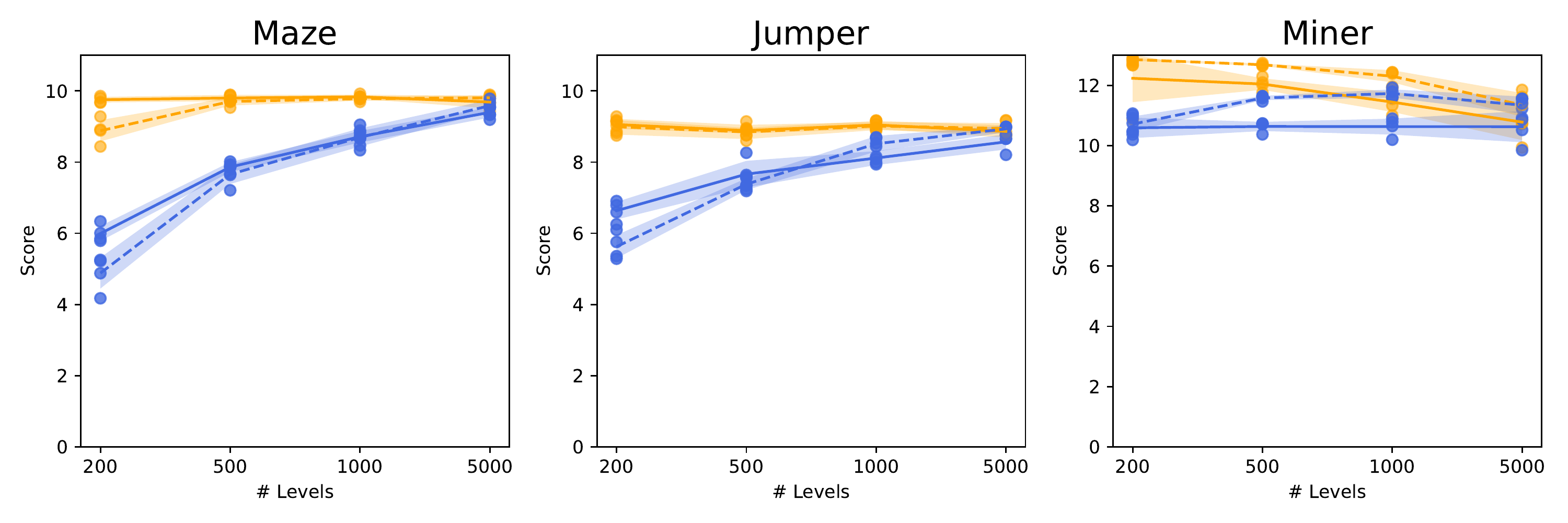}
         \caption{Extrinsic Reward}
         \label{fig:maxEnt_extReward_reward_withRNN}
     \end{subfigure}
    \caption{\textbf{Generalization ability of maximum entropy and extrinsic reward policy:} 
    (\textbf{a.top row}) Score of maximum entropy policy, normalized by the oracle score.
    (\textbf{a.bottom row}) Success Rate of maximum entropy policy.
    (\textbf{b}) Score of extrinsic reward policy. 
    Training for maximum entropy exhibits a small generalization gap in Maze, Jumper and Miner. Average and standard deviation are obtained using $4$ seeds.}
    \label{fig:maxEnt_extReward_wothRNN}
\end{figure}

In Fig.~\ref{fig:maxEnt_extReward_entropy_withRNN} the top row describes the intrinsic return of the maxEnt policy, normalized by the oracle's return (described above). We draw a comparison between agents with memory (GRU) and without. For the Maze environment (top-right), the agent achieves over $90\%$ of the oracle's performance when employing a memory unit (GRU). For the Jumper and Miner environments, the agents approach $60\%$ and $80\%$ of the oracle's score, respectively. 

Next, the bottom row of Fig.~\ref{fig:maxEnt_extReward_entropy_withRNN} details the success rate as the ratio of instances in which the agent successfully reached the oracle's score (meaning that the entire traversable region has been explored). For the Maze environment, the agent achieves a success-rate of $70\%$, whereas for the Jumper and Miner the agent fails to meet the oracle's score. This indicates that the oracle, as per our implementation for Jumper and Miner, captures states that are in-effect unreachable, and thus the agent is unable to match the oracle's score.

\subsection{The Importance of a Memory Unit for maxEnt}
\label{appendix:maxent_memory}

We evaluate the importance of memory for maxEnt: In Fig.~\ref{fig:maxEnt_extReward_entropy_withRNN} we train a maxEnt agent with and without memory (GRU). The results indicate that for Maze, the GRU is vital in order to maximize performance for all various sizes of training set. In Jumper, we see that the GRU provides an advantage when training on $200$  levels, however the benefit diminishes when additional training levels are available. For Miner, there appears to be no significant advantage for incorporating memory with $200$ training levels. However, the benefit of a GRU becomes noticeable once more training levels are available (beyond $500$).

When looking at the extrinsic reward (Fig.~\ref{fig:maxEnt_extReward_reward_withRNN}), we see an interesting effect after introducing a GRU. Maze and Jumper suffer a degradation in performance with $200$ levels (indicating overfit), while Miner appears to be unaffected.

In summary, we empirically show that memory is beneficial for the maxEnt policy on Maze, Jumper and Miner. Interestingly, we demonstrate that the introduction of memory for training to maximize extrinsic reward causes the agent to overfit in Maze and Jumper with $200$ training levels.

\section{Experimental Setup}
\label{appendix:setup}
This section describes the constants and hyperparameters used as part of the evaluation of our algorithm.
\subsection{Normalization Constants} \label{appendix:norm_constants}
In Figures~\ref{fig:Test_performance_on_Procgen} and~\ref{fig:Test_performance_on_Procgen_idaac_expgen}  we compare the performance of the various algorithms. The results are normalized in accordance with \citep{cobbe2020leveraging}, which defines the normalized return as $$R_{norm} = (R -R_{min})/(R_{max} - R_{min}).$$ The $R_{min}$ and $R_{max}$ of each environment are detailed in Table~\ref{table:normalization_constants}. Note that the test performance of \ppo{} on Heist   
(see Fig.~\ref{fig:Test_performance_on_Procgen}) is lower than $R_{min}$ (the trivial performance), indicating severe overfitting.

\begin{table*}
\begin{center}
 \begin{tabular}{||c c c c c||}
 \multicolumn{1}{c}{} & \multicolumn{2}{c}{Hard} & \multicolumn{2}{c}{Easy} \\
 \hline
 Environment & $R_{min}$ & $R_{max}$ & $R_{min}$ & $R_{max}$ \\
 \hline\hline
 CoinRun & 5 & 10 & 5 & 10 \\ 
 StarPilot & 1.5 & 35 & 2.5 & 64\\ 
 CaveFlyer & 2 & 13.4 & 3.5 & 12 \\ 
 Dodgeball & 1.5 & 19 & 1.5 & 19 \\ 
 FruitBot & -.5 & 27.2 & -1.5 & 32.4 \\ 
 Chaser & .5 & 14.2 & .5 & 13 \\ 
 Miner & 1.5 & 20 & 1.5 & 13 \\ 
 Jumper & 1 & 10 & 3 & 10 \\ 
 Leaper & 1.5 & 10 & 3 & 10 \\ 
 Maze & 4 & 10 & 5 & 10 \\ 
 BigFish & 0 & 40 & 1 & 40 \\ 
 Heist & 2 & 10 & 3.5 & 10 \\ 
 Climber & 1 & 12.6 & 2 & 12.6 \\ 
 Plunder & 3 & 30 & 4.5 & 30 \\ 
 Ninja & 2 & 10 & 3.5 & 10 \\ 
 BossFight & .5 & 13 & .5 & 13 \\ 
 \hline
 \end{tabular}
 \caption{Normalization Constants.}
 \label{table:normalization_constants}
 \end{center}
\end{table*}

\begin{table}[ht]
\centering
\begin{tabular}{ |p{3cm}||c| }
  \hline
 Game     &   $k$     \\
 \hline
Maze & $6$ \\
Jumper & $4$ \\
Miner & $2$ \\
Heist & $8$ \\
BigFish & $8$ \\
StarPilot & $1$ \\
FruitBot & $1$ \\
BossFight & $1$ \\
Plunder & $2$ \\
CaveFlyer & $2$ \\
CoinRun & $1$ \\
Chaser & $2$ \\
Climber & $2$ \\
Dodgeball & $2$ \\
Leaper & $1$ \\
Ninja & $2$ \\
 \hline
\end{tabular}
\vskip 0.08in
\caption{Consensus size $k$ as hyperparameter for each game.}
\label{table:consensus_k}
\end{table}

\begin{table}[!ht]
\centering
\begin{tabular}{ |p{2cm}||p{2cm}|p{2cm}|p{2cm}|p{2cm}|  }
 \hline
Ensemble size    &   4    &   6  &  8 & 10 \\
 \hline
\methodnameshort   & $8.02\pm0.06$ & $8.15\pm0.19 $& $8.00\pm0.12$ &$\pmb{8.22\pm0.11}$ \\
 \hline
\end{tabular}
\vskip 0.08in
\caption{Ablation study of ensemble size and its effect on the test score. Each network in the ensemble is trained on $200$ instances of Maze. The results show improved performance for large ensemble size. The mean and standard deviation are computed using $10$ runs with different seeds.}
\label{table:ensemble_size_maze}
\end{table}

\begin{table}[ht]
\centering
\begin{tabular}{ |p{5cm}||c| }
  \hline
Parameter & Value \\
 \hline
$\gamma$   & .999 \\ 
$\lambda$  & .95 \\
\# timesteps per rollout  & 512 \\
Epochs per rollout  & 3 \\
\# minibatches per epoch  & 8 \\ 
Entropy bonus ($k_H$)  & .01 \\
PPO clip range  & .2 \\
Reward Normalization?  & Yes \\
Learning rate  & 5e-4 \\
\# workers  & 1 \\
\# environments per worker  & 32 \\ 
Total timesteps  & 25M \\
GRU?  & Only for maxEnt \\
Frame Stack?  & No \\
 \hline
\end{tabular}
\vskip 0.08in
\caption{PPO Hyperparameters.}
\label{table:Hyperparameters}
\end{table}

\subsection{Hyperparameters} \label{appendix:hyperparameters}
As described in Section~\ref{sec:algorithm}, the hyperparameters of our algorithm are the number of agents $m$ that form the ensemble, of which $k$ agents are required to be in agreement for the ensemble to achieve a consensus on its action, and $\alpha$ as the parameter of $n_{\pi_{\ent}}\sim Geom(\alpha)$ that represents the number of maxEnt steps taken when the ensemble fails to reach a consensus. An additional hyperparameter is the neighborhood size $k_{\mathrm{NN}}$ of the $k$-NN estimator (Section~\ref{subsection:training_maxEnt}), used to determine the reward of the maxEnt policy.

We conducted a hyperparameter search over the ensemble size $m \in \{4,6,8,10\}$ for different values of ensemble agreement $k \in \{2,4,6,8\}$, and values of $\alpha \in \{0.2, 0.5, 0.8\}$. We found that a value of $\alpha=0.5$, and an ensemble size of $m=10$ produce the best results for all games, whereas the value of $k$ varies from game to game, as detailed in table~\ref{table:consensus_k}. For example, table~\ref{table:ensemble_size_maze} shows the results for varying values of $m$ and $k$ for the Maze environment. 
Throughout our experiments, we train our networks using the Adam optimizer~\citep{kingma2014adam}. For the \ppo~hyperparameters we use the hyperparameters found in~\citep{cobbe2020leveraging} as detailed in Table~\ref{table:Hyperparameters}. 

Table~\ref{tab:results_for_diff_k_train_maze} shows an evaluation of the maxEnt gap and \methodnameshort~using different neighbor size $k_{\mathrm{NN}}$ on the Maze environment. We found that the best performance is obtained for $k_{\mathrm{NN}} \in \{1, 2, 3, 4, 5\}$. Thus, we choose $k_{\mathrm{NN}} = 2$, the second nearest neighbor, for all games.

\begin{table*}[ht]
    \small
    \centering
    \begin{tabular}{|c|cccc|}
    \toprule
    Neighbor &  & Maze & Environment &  \\
     Size $k_{\mathrm{NN}}$  &  maxEnt (Train) &  maxEnt (Test)  &  maxEnt Gap [\%]& ExpGen (Test)  \\
    \toprule
    1 & ${17.5\pm 0.5}$  &	${15.3\pm 1.4}$   & $\pmb{12.7\%}$ & $7.9\pm 0.2$ \\
    2 & ${33.9\pm 0.2}$ & ${31.3\pm 1.5}$ &  $\pmb{7.7\%}$ & $8.3\pm 0.2$  \\
    3 & ${50.2\pm 2.0}$ & ${42.5\pm 2.8 }$ &  $\pmb{15.3\%}$ & $8.2\pm 0.2$  \\
    4 & ${62.9\pm 2.6}$ & ${52.3\pm 3.0 }$ &  $\pmb{16.8\%}$ & $8.2\pm 0.1$  \\
    5 & ${70.6\pm 3.3}$ & ${57.8\pm 4.6 }$ &  $\pmb{15.5\%}$ & $8.2\pm 0.1$  \\
    6 & ${80.0\pm 3.2}$ & ${65.3\pm 1.4 }$ &  ${18.4\%}$ & $8.1\pm 0.2$  \\
    7 & ${87.2\pm 1.4}$ & ${71.1\pm 1.1 }$ &  ${18.5\%}$ & $8.0\pm 0.1$  \\
    8 & ${93.8\pm 1.0}$ & ${75.3\pm 2.3 }$ &  ${19.7\%}$ & $7.8\pm 0.2$  \\
    9 & ${96.6\pm 4.8}$ & ${78.0\pm 2.0 }$ &  ${19.3\%}$ & $8.1\pm 0.2$  \\
    10 & ${102.2\pm 1.8}$ & ${89.1\pm 0.6 }$ &  $\pmb{12.8\%}$ & $7.9\pm 0.1$  \\
    \bottomrule
    \end{tabular}
    \caption{Hyperparameter search for neighborhood size $k_{\mathrm{NN}}$ for the Maze environment. The table presents the maxEnt gap and the performance of ExpGen for varying values of $k_{\mathrm{NN}}$. The mean and variance are computed for $3$ seeds.}
    \label{tab:results_for_diff_k_train_maze}
\end{table*}

% TODO: fill-in the results for Heist
% \begin{table*}[ht]
%     \small
%     \centering
%     \begin{tabular}{c|cccc|}
%     \toprule
%     Neighbor &  maxEnt (Train) &  maxEnt (Test)  &  maxEnt Gap [\%]& ExpGen (Test)  \\
%      Size $k_{\mathrm{NN}}$  &   &   &  &  \\
%     \toprule
%     1 & $\pmb{17.854\pm 0.}$  &	$\pmb{14.869\pm 0.}$   & $\pmb{16.7\%}$ & $7.0\pm 0.5$ \\
%     2 & $\pmb{35.279\pm 0.}$ & $\pmb{30.241\pm 0. }$ &  $\pmb{14.3\%}$ & $6.9\pm 0.4$  \\
%     3 & $\pmb{50.52\pm 0.}$ & $\pmb{43.507\pm 0. }$ &  $\pmb{13.9\%}$ & $8.2\pm 0.2$  \\
%     4 & $\pmb{60.042\pm 0.}$ & $\pmb{54.539\pm 0. }$ &  $\pmb{9.2\%}$ & $8.2\pm 0.1$  \\
%     5 & $\pmb{70.309\pm 0.}$ & $\pmb{59.39\pm 0. }$ &  $\pmb{15.5\%}$ & $8.2\pm 0.1$  \\
%     6 & $\pmb{77.805\pm 0.}$ & $\pmb{66.337\pm 0. }$ &  $\pmb{14.7\%}$ & $8.1\pm 0.2$  \\
%     7 & $\pmb{91.144\pm 0.}$ & $\pmb{73.117\pm 0. }$ &  $\pmb{19.7\%}$ & $8.0\pm 0.1$  \\
%     8 & $\pmb{97.305\pm 0.}$ & $\pmb{77.777\pm 0. }$ &  $\pmb{20.1\%}$ & $7.8\pm 0.2$  \\
%     9 & $\pmb{102.885\pm 0.}$ & $\pmb{82.023\pm 0. }$ &  $\pmb{20.3\%}$ & $8.1\pm 0.2$  \\
%     10 & $\pmb{102.262\pm 0.}$ & $\pmb{85.947\pm 0. }$ &  $\pmb{16.0\%}$ & $7.9\pm 0.1$  \\
%     \bottomrule
%     \end{tabular}
%     \caption{\NEW{Hyperparameter search for neighborhood size $k_{\mathrm{NN}}$ for the Heist environment. The table presents the maxEnt gap and the performance of ExpGen for varying values of $k_{\mathrm{NN}}$. The mean and variance are computed for $3$ seeds.}}
%     \label{tab:results_for_diff_k_train_heist}
% \end{table*}

\section{Results for all ProcGen Games} \label{appendix:results_all_idaac_procgen}
\begin{figure}[ht]
\vspace{-1cm}
\includegraphics[width=.9\textwidth]{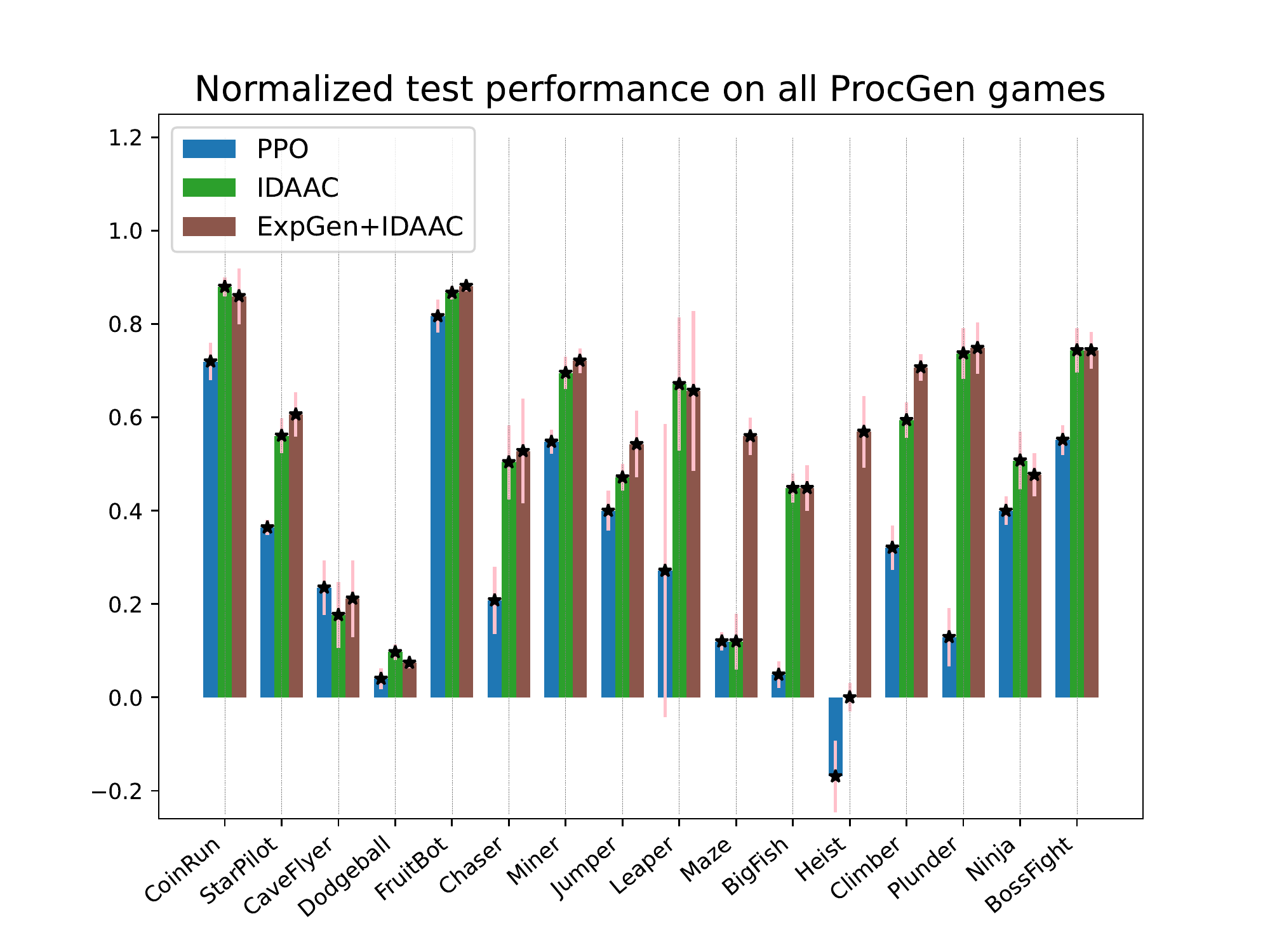}
\caption{
Normalized test Performance for \ppo{}, \idaac{}, and \methodnameshort+\idaac{}, on all ProcGen games. \methodnameshort~achieves state-of-the-art performance on test levels of Maze, Heist, and Jumper and on-par performance in the remaining games.  }
\label{fig:Test_performance_on_Procgen_idaac_expgen}
\end{figure}

Figure~\ref{fig:Test_performance_on_Procgen_idaac_expgen} details the normalized test performance for all ProcGen games. Normalization is performed according to \citep{cobbe2020leveraging} as described in Appendix~\ref{appendix:norm_constants}. The figure demonstrates that \methodnameshort~establishes state-of-the-art results on several challenging games and achieves on-par performance with the leading approach on the remaining games.

\section{Results after convergence} \label{appendix:table_50}
Tables \ref{table:train_det_nondet50} and \ref{table:test_det_nondet50} detail the train and test performance of \methodnameshort, \leep{} and \ppo{}, when trained for $50M$ environment steps. Table \ref{table:test_det_nondet50} shows that \methodnameshort~surpasses \leep{}~and \ppo{}~in most games. 

\begin{table*}[ht]
\centering
\begin{tabular}{ |p{3cm}||p{2cm}|p{2cm}|p{4cm}|  }
 \hline
 Game          &   PPO           &    LEEP        &  \methodnameshort \\
 \hline
 Maze      & $9.45 \pm 0.21$ & $9.84 \pm 0.05$ & $9.71 \pm 0.11 $\\
 Heist     & $7.97 \pm 0.56$ & $6.86 \pm 0.68 $& $9.62\pm 0.08 $\\
 Jumper    & $8.62 \pm 0.08$ & $6.1\pm0.4$ & $8.00 \pm 0.13 $\\
 Miner     & $12.86 \pm 0.06$ & $1.9\pm0.2$ & $12.53 \pm 0.12 $\\
 BigFish   & $14.24 \pm 3.36$ & $8.82 \pm 0.36$ & $5.95 \pm 0.32 $\\
 Climber   & $8.76 \pm 0.41$ & $4.6\pm0.4$     & $ 8.89 \pm 0.29 $\\
 Dodgeball & $8.88\pm 0.38$ & $6.22 \pm 0.55$ & $8.73 \pm  0.40 $\\ 
 Plunder   & $9.64\pm 1.85$ & $5.1\pm0.1$ & $8.06 \pm 0.98 $\\
 Ninja     & $9.10\pm 0.32$ & $5.2\pm0.3$ & $8.90 \pm 0.30 $\\
 CaveFlyer & $8.98 \pm 0.59$ & $5.4\pm0.1$ & $8.75 \pm 0.59 $\\
 \hline
\end{tabular}
\caption{\textbf{Train score} of ProcGen environments trained on 200 instances for 50M environment steps. We compare our algorithm to the baselines LEEP and PPO. The mean and standard deviation are computed over 8 runs with different seeds.}
\label{table:train_det_nondet50}
\end{table*}

\begin{table*}[ht]
\centering
\begin{tabular}{ |p{3cm}||p{2cm}|p{2cm}|p{4cm}|  }
 \hline
 Game          &   PPO           &    LEEP         &  \methodnameshort \\
 \hline
 Maze      & $5.78 \pm 0.39$ & $6.78 \pm 0.21$ & $\pmb{8.33 \pm 0.14} $\\
 Heist     & $2.54 \pm 0.45$ & $4.42 \pm 0.57$& $\pmb{6.91 \pm 0.24}$\\
 Jumper    & $5.78 \pm 0.28$ &  $6.4\pm0.4$  & $\pmb{6.64 \pm 0.15} $\\
 Miner     & $8.76 \pm 0.33$ &  $0.8\pm0.1$   & $\pmb{9.48 \pm 0.39} $\\
 BigFish   & $3.82 \pm 1.98$ & $5.5 \pm 0.41$ & $ \pmb{5.99 \pm 0.64} $\\
 Climber   & $6.14 \pm 0.50$ &  $2.6\pm0.4$   & $6.29\pm 0.54 $\\
 Dodgeball & $3.71 \pm 0.55$ & $\pmb{4.58 \pm 0.47}$& $3.84 \pm 0.56$\\ 
 Plunder   & $\pmb{7.78 \pm 1.74}$ & $4.2\pm0.2$ & $6.91 \pm 1.00 $\\
 Ninja     & $6.94 \pm 0.30$ & $4.9\pm0.8$ & $\pmb{6.75 \pm 6.75} $\\
 CaveFlyer & $6.19 \pm 0.66$ & $2.6\pm0.2$ & $\pmb{6.36 \pm 0.49} $\\

 \hline
\end{tabular}
\caption{\textbf{Test score} of ProcGen environments trained on $200$ instances for $50M$ environment steps. We compare our algorithm to the baselines LEEP and PPO. The mean and standard deviation are computed over $8$ runs with different seeds.}
\label{table:test_det_nondet50}
\end{table*}

\section{Ablation Study}\label{appendix:ablation_study}

In the following sections, we provide ablation studies of an \methodnameshort~variant that combines random actions and an evaluation of $L_0$ state similarity measure.

\subsection{Ensemble Combined with Random Actions}
We compare the proposed approach to a variant of \methodnameshort~denoted by Ensemble+random where we train an ensemble and at test time select a random action if the ensemble networks fail to reach a consensus. The results are shown in Table \ref{table:ablation_random}, indicating that selecting the maximum entropy policy upon ensemble disagreements yields superior results.

\begin{table}[ht]
\centering
\begin{tabular}{ |p{3cm}||c|c| }
  \hline
 Algorithm     &   Train & Test     \\
 \hline
 \methodnameshort    &$\pmb{9.6\pm0.2}$& $\pmb{8.2 \pm 0.1}$ \\
 Ensemble + random & $9.3\pm0.1$ & $6.2 \pm 0.2$ \\
 LEEP         &$9.4\pm0.3$& $6.6 \pm 0.2$ \\
 PPO + GRU    & $9.5\pm0.2$& $5.4 \pm 0.3$ \\
 PPO          &$9.1\pm0.2$ & $5.6 \pm 0.1$\\
 \hline
\end{tabular}
\vskip 0.08in
\caption{Ablation study of \methodnameshort~on Maze. The table shows testing scores of networks trained on $200$ maze instances. We present a comparison between the proposed approach and LEEP, PPO+GRU and PPO, as well as an alternative ensemble policy with random actions upon ensemble disagreement. The mean and standard deviation are computed using $10$ runs with different seeds.}
\label{table:ablation_random}
\end{table}

\subsection{Evaluation of Various Similarity Measures for maxEnt}
\label{appendix:L0_L2_norms}
Tables~\ref{tab:results_L0_L2_train} and~\ref{tab:results_L0_L2_test} present the results of our evaluation of \methodnameshort~equipped with a maxEnt exploration policy that uses either the $L_0$ or $L_2$ norms. The experiment targets the Maze and Heist environments and uses the same train and test procedures as in the main paper ($25M$ training steps, score mean and standard deviation are measured over $10$ seeds).

\begin{table*}[ht]
    \small
    \centering
    \begin{tabular}{l|c|c|c}
    \toprule
    Game & ExpGen $L_0$ (Train) &  ExpGen $L_2$ (Train)  & PPO (Train)  \\
    \toprule
    Heist&	    $\pmb{9.4\pm 0.3}$    & $\pmb{9.4\pm 0.1}$   &$6.1\pm 0.8$ \\
    Maze&	   $\pmb{9.6\pm 0.2}$   &	$\pmb{9.6\pm 0.1}$  & $9.1\pm 0.2$   \\
    \bottomrule
    \end{tabular}
    \caption{Train scores of \methodnameshort~using maxEnt policy with either $L_0$ or $L_2$ compared with \ppo{}. The mean and standard deviation are measured over $10$ seeds.}
    \label{tab:results_L0_L2_train}
\end{table*}

\begin{table*}[ht]
    \small
    \centering
    \begin{tabular}{l|c|c|c}
    \toprule
    Game &  ExpGen $L_0$ (Test) &  ExpGen $L_2$ (Test) &  PPO (Test) \\
    \toprule
    Heist&	    $\pmb{7.4\pm 0.1}$  &	$\pmb{7.4\pm 0.2}$   & $2.4\pm 0.5$  \\
    Maze&	   $\pmb{8.2\pm 0.1}$ &	$\pmb{8.3\pm 0.2}$ &  $5.6\pm 0.1$   \\
    \bottomrule
    \end{tabular}
    \caption{Test scores of \methodnameshort~using maxEnt policy with either $L_0$ or $L_2$ compared with \ppo{}. The mean and standard deviation are measured over $10$ seeds.}
    \label{tab:results_L0_L2_test}
\end{table*}

The results demonstrate that both $L_2$ and $L_0$ allow \methodnameshort~to surpass the \ppo{}~baseline for Maze and Heist environments, in which they perform similarly well at test time. This indicates that both are valid measures of state similarity for the maxEnt policy.

\section{Sample Complexity}
\begin{figure}[ht]
    \centering
     \begin{subfigure}[h]{0.65\textwidth}
        \centering
         \includegraphics[width=\textwidth]{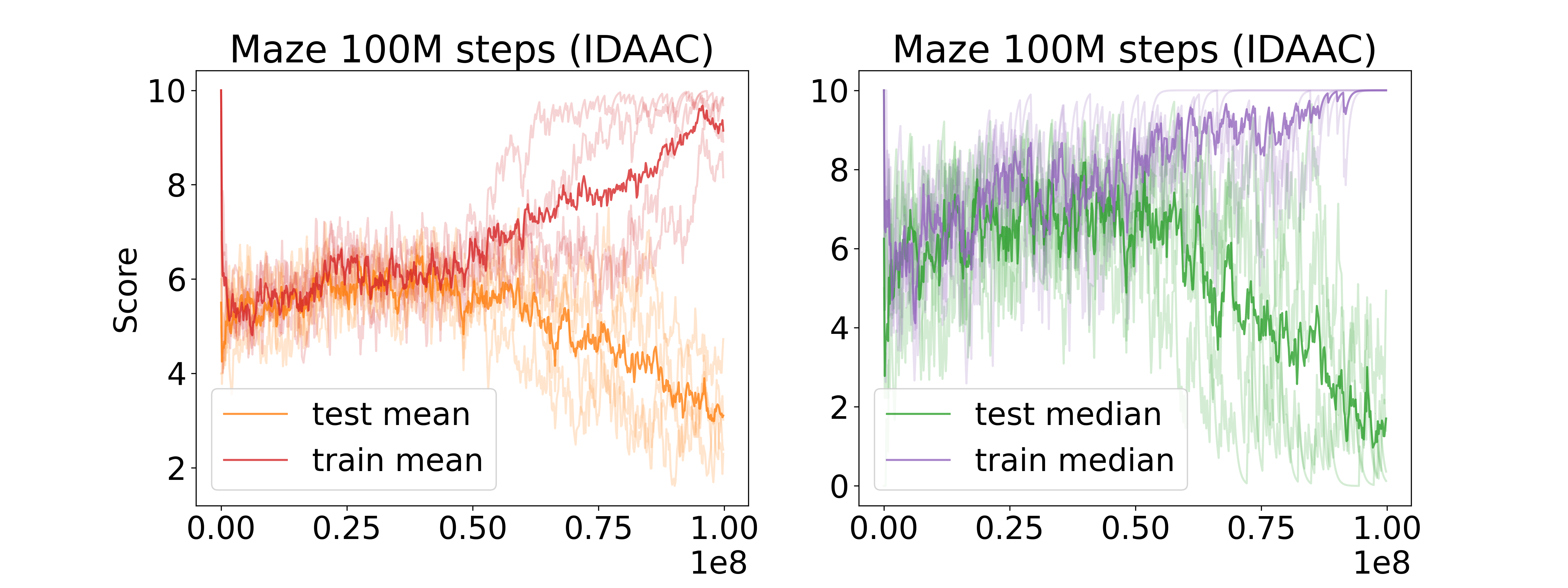}
         \label{idaac_Maze_100M}
     \end{subfigure}
     \begin{subfigure}[h]{0.65\textwidth}
        \centering
         \includegraphics[width=\textwidth]{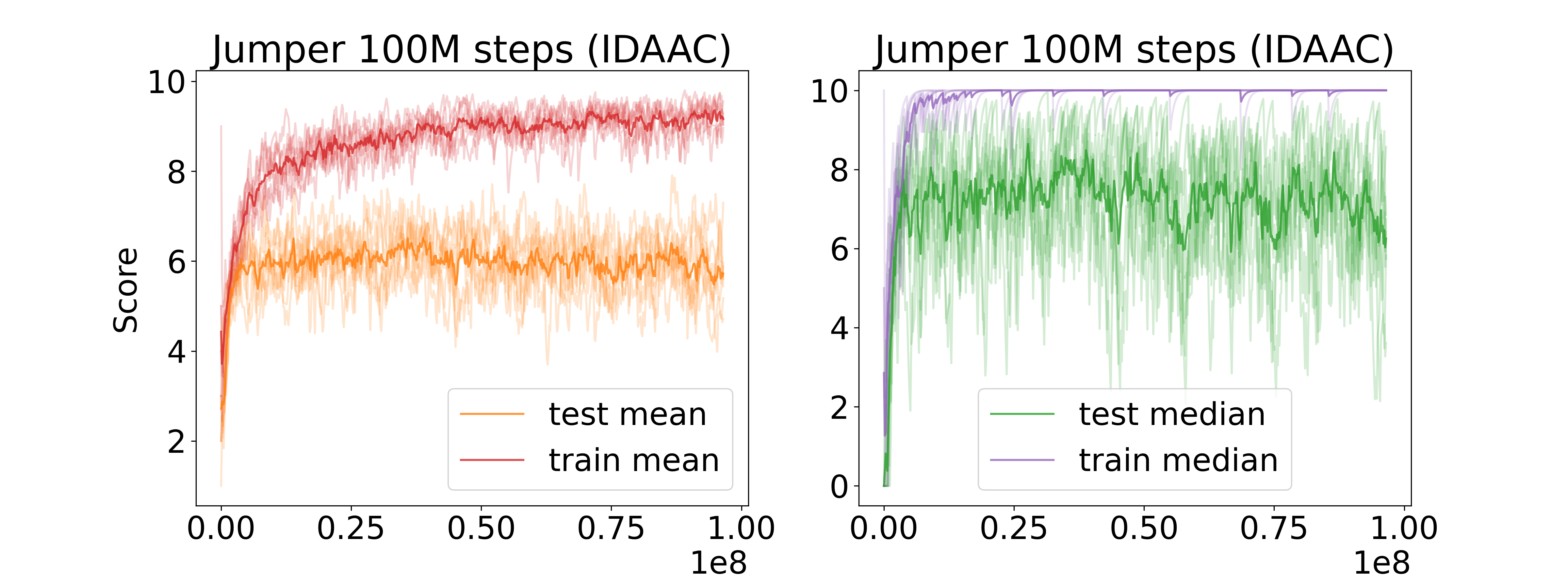}
         \label{idaac_Jumper_100M}
     \end{subfigure}
    \caption{The mean and median of the accumulated reward for \idaac{}~trained for $100M$ steps, averaged over $10$ runs with different seeds. The curves show that test-reward stagnates and even decreases beyond $25M$ steps.} 
    \label{fig:idaac_100Msteps}
\end{figure}

One may wonder whether the leading approaches would benefit from training on additional environment steps.
A trained agent can still fail at test time either due to poor generalization performance (overfitting on a small number of training domains) or due to insufficient training steps of the policy (underfitting). In this work, we are interested in the former and design our experiments such that no method underfits.
Figure~\ref{fig:idaac_100Msteps} shows \idaac{}~training for $100M$ steps on Maze and Jumper, illustrating that the best test performance is obtained at around $25M$ steps, and training for longer does not contribute further (and can even degrade performance). Therefore, although \methodnameshort~requires more environment steps (on account of training its ensemble of constituent reward policies), training for longer does not place our baseline (\idaac{}) at any sort of a disadvantage.

\newpage

\end{document}